\documentclass[11pt]{article}

\usepackage{hyperref}
\usepackage{makecell}
\usepackage{float}
\usepackage{amsmath}
\usepackage{color}
\usepackage{amssymb}
\usepackage[round]{natbib}
\usepackage{caption}
\usepackage{subcaption}
\usepackage{inputenc}
\usepackage[export]{adjustbox}
\usepackage{times}
\usepackage{graphicx}
\usepackage{lineno}
\usepackage{enumitem}

\newcommand{\toprule}{\hline}
\newcommand{\midrule}{\hline}
\newcommand{\bottomrule}{\hline}
\newcommand{\new}[1]{{#1}}

\title{\vspace{-35pt}Scalable Semantic 3D Mapping of Coral Reefs \\with Deep Learning\vspace{-7pt}
}

\author
{Jonathan Sauder,$^{1, 2, \ast}$ Guilhem Banc-Prandi,$^{2}$ Anders Meibom$^{2, 3}$, Devis Tuia$^{1}$\\
	\scriptsize{$^{1}$Environmental Computational Science and Earth Observation Laboratory, École Polytechnique Fédérale de Lausanne, CH-1950 Lausanne, Switzerland}\\
	\scriptsize{$^{2}$Laboratory for Biological Geochemistry, École Polytechnique Fédérale de Lausanne, CH-1015 
		Lausanne, Switzerland}\\
	\scriptsize{$^{3}$Center for Advanced Surface Analysis, University of Lausanne, CH-1015 
		Lausanne, Switzerland}\\
	\scriptsize{$^\ast$To whom correspondence should be addressed; E-mail:  jonathan.sauder@epfl.ch.\vspace{-17pt}}
}

\date{}

\begin{document}

	\baselineskip16pt

	\maketitle

	
	
	
	\begin{abstract}{} \normalsize 
		Coral reefs are among the most diverse ecosystems on our planet, and essential to the livelihood of hundreds of millions of people who depend on them for food security, income from tourism and coastal protection. Unfortunately, most coral reefs are existentially threatened by global climate change and local anthropogenic pressures. To better understand the dynamics underlying deterioration of reefs, monitoring at high spatial and temporal resolution is key. However, conventional monitoring methods for quantifying coral cover and species abundance are limited in scale due to the extensive manual labor required. 
		Although computer vision tools have been employed to aid in this process, in particular Structure-from-Motion (SfM) photogrammetry for 3D mapping and deep neural networks for image segmentation, \new{analysis of the data products creates a bottleneck, effectively limiting their scalability.}\\ \vspace{-9pt}
		
		This paper presents a new paradigm for mapping underwater environments from ego-motion video, unifying 3D mapping systems that use machine learning to adapt to challenging conditions under water, combined with a modern approach for semantic segmentation of images.\\ \vspace{-9pt}
		
		The method is exemplified on coral reefs in the northern Gulf of Aqaba, Red Sea, demonstrating high-precision 3D semantic mapping at unprecedented scale with significantly reduced required \new{labor costs: a 100 m video transect acquired within 5 minutes of diving with a cheap consumer-grade camera can be fully automatically transformed into a semantic point cloud within 5 minutes.} 
		\new{We demonstrate the spatial accuracy of our method and the semantic segmentation performance, and publish a large dataset of ego-motion videos from the northern Gulf of Aqaba, along with a dataset of video frames annotated for dense semantic segmentation of benthic classes.}\\\vspace{-9pt}
		
		Our approach significantly scales up \new{coral reef monitoring by taking a leap towards fully automatic analysis of video transects. The method democratizes coral reef transects by reducing the labor, equipment, logistics, and computing cost. This can} help to inform conservation policies \new{more efficiently. The underlying computational method of learning-based Structure-from-Motion has} broad implications for \new{fast low-cost} mapping of underwater environments other than coral reefs.
		
	\end{abstract}
	
	\centerline{\textbf{Keywords:} Artificial Intelligence, Computer Vision, Coral Ecology, Coral Reefs, Machine Learning, }
	\centerline{\new{Structure From Motion, Monocular Depth Estimation, Visual Odometry, Semantic Segmentation, 3D Vision}}
	\vfill
	\clearpage
	\section*{Introduction}
	\vspace{2pt}
	
	Coral reefs are among the most diverse ecosystems on the planet: despite covering less than 0.1\% of the planet’s surface area, they host at least 32\% of known marine species \citep{blabla}. Up to half a billion people worldwide rely on the services provided by coral reefs, which include food security and tourism \citep{halfabillion}.
	\new{Coral reefs are at a decline worldwide \cite{icrireport}, as they locally suffer from detrimental human activities, and are globally threatened by increasingly warm oceans, which can cause corals to bleach and eventually die \citep{hughes2017global, knowlton2008shifting}.}
	
	\new{Under current greenhouse gas emission trajectories \citep{ipcc21}}, almost all warm-water coral reefs are projected to suffer significant losses of area or local extinction, even if global warming is limited to 1.5°C \citep{ipcc15}.
	Therefore, coral reefs are among the ecosystems that are most vulnerable to climate change. The frequency of mass bleaching events, in which vast areas of reefs bleach at once, will increase in the future \citep{future}, giving many reefs little hope to recover in between.
	\vspace{2pt}
	
	However, there is an extremely high variability in resilience to stresses between various regions, species, and even genotypes of the same species. Remarkably, regions with reefs that could withstand end-of-century ocean temperatures and acidity have been identified \citep{beyer2018risk}. For example, in the Gulf of Aqaba in the northern Red Sea, prominent species of corals exhibit an exceptionally high thermal tolerance, promising to withstand sea temperature increase of more than 5°C \new{\citep{maoz, voolstra, kruegerredsea, osmanredsea, savaryredsea}}. 
	Therefore, for corals in such refugia, the most imminent threats are local stresses, caused by destructive fishing practices, overtourism, urbanization of the coastlines, and associated local pollution. To ensure the survival of coral reefs until the end of the century and beyond, it is imperative to get a better understanding of the dynamics of how global temperature rise and local anthropogenic pollution damage coral reefs and evaluate if and how the reefs recover from them. This necessitates \new{efficient} methods for large scale monitoring \new{at high spatial and temporal resolution.}
	\vspace{2pt}
	
	Conventional methods for monitoring corals are often inadequate in terms of scalability because they are highly labor intensive. The most universally recognized and applied method consists of line transects \& photo quadrats \citep{conventional, icriconventional}, in which photos of the seafloor inside a quadratic frame of a known reference size are taken along a straight transect  line.
	Experts then analyze the photos, determining the presence of species and their health status, and subsequently extrapolate to larger areas. The results \new{can be} heavily biased by the exact location\new{s} of the \new{quadrats}, details in the photo quadrat protocol, and the human analysts processing the photos. This makes a direct comparison across studies challenging \citep{icrireport}, with scientists essentially only agreeing on coarse metrics, such as the percentage of live coral cover. Most importantly, the involved manual effort of accurately managing and analyzing the photo quadrats, even for a relatively small transect, is very large. \new{Nonetheless, transect lines can be rapidly deployed by divers and the GPS coordinates of the end-points accurately determined by means of marker buoys. These logistical consideration make data acquisition along transects the de-facto standard for coral reef monitoring. To collect data even faster and with less logistical overhead than photo quadrats, video transects \citep{videotransect} can also be acquired. However, determining benthic cover from video is even more cumbersome for analysts due to lack of normalized reference objects and possible overlaps between video frames.} Coral monitoring tools of the future absolutely must automate the labor-intensive process \new{of analyzing transect data.}
	\vspace{2pt}
	
	A vast proportion of coral reefs are found in countries with restricted research resources. To date, coral reef monitoring efforts have heavily focused on accessible reefs in wealthy countries. One key aspect for scalable monitoring tools is that \new{their costs in terms of required human resources to analyze the data and in terms of logistics of diving operations should be as low as possible. Furthermore,} the required equipment should be available worldwide \new{within a reasonable budget}. While hyperspectral sensors can facilitate identification of corals \citep{hyperdiver, digitizing, airborne}, their price can be prohibitive for widespread use. On the other hand, \new{the cost of underwater color cameras has dramatically fallen, which suggests} computer vision \new{tools can be applied to successfully scale automated semantic segmentation of living corals from color camera imagery.}
	\vspace{2pt}

	\begin{figure}[t]
		\centering
		\begin{tabular}{p{6.7cm}p{8cm}}
			\multicolumn{2}{c}{\includegraphics[width=\linewidth]{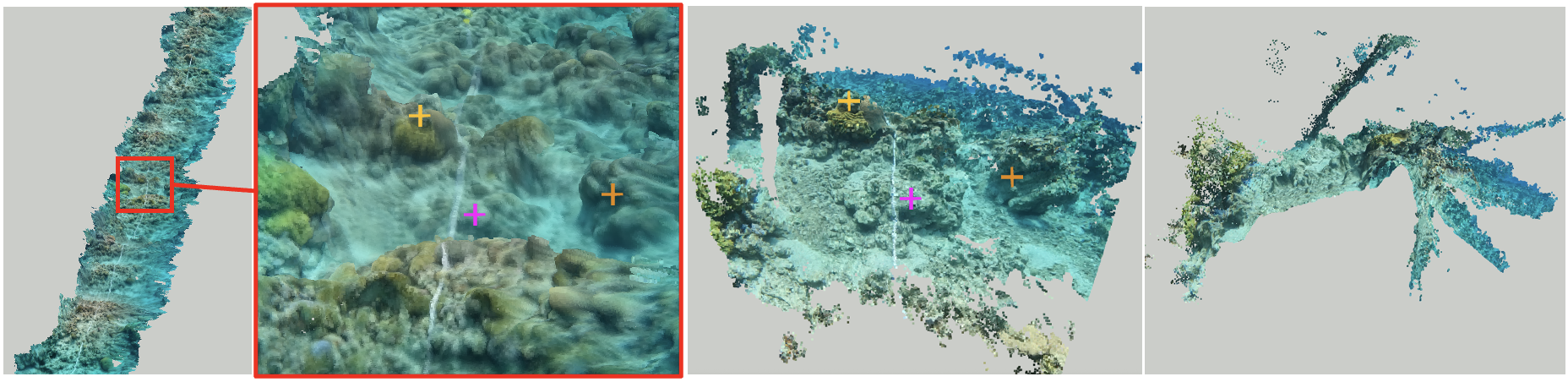}}\\
			a) Our method & b) Metashape \\
		\end{tabular}
		\caption{
			Existing conventional SfM fails to produce a coherent point cloud from uncurated image collections such as video frames. \new{This example shows the point clouds from a video transect in the King Abdullah Reef in Aqaba, Jordan. Leftmost panel: our proposed method creates a coherent point cloud for the whole transect in 310 seconds. Rightmost panel: Agisoft Metashape \citep{agisoft} fails to capture a globally coherent structure and only produces a point cloud of a small section of the transect (aligning $69$ out of $1889$ images) after 2 hours and 32 minutes, albeit at higher resolution than our method for the part successfully reconstructed. The colorful markers in the two zoomed in versions (center panels) show the same spatial features in the two point clouds.}}
		\label{fig:failsfm}
	\end{figure}

	While many computer vision tools have been proposed to aid in coral reef monitoring, the \new{scalability in terms of fully analyzed transects per time unit and money unit} remains limited. The main lines of \new{computer vision} work can be summarized into computer-aided mapping, commonly realized with Structure-from-Motion (SfM) photogrammetry \new{\citep{burns, leon, storlazzi, alonso2019coralseg, reefscapegenomics, sfm1, reverse, orthomosaic}}, and benthic cover analysis systems, which use machine learning to recognize coral types and other benthic classes in images \citep{coralnet, svmcoral, coralnetvsreal, coralnetengine, digitizing}. However, underwater environments pose particular challenges to computer vision methods due to difficult lighting conditions and diffraction effects, caustics, non-linear attenuation and scenes with many dynamic objects. This implies that computer vision algorithms often only work reliably under controlled conditions. SfM mapping techniques, for example, are often brittle. They require carefully curated high-resolution image collections for 3D reconstruction or they will fail to create coherent 3D models, as shown in Figure~\ref{fig:failsfm}.
	At the same time, they are limited in scale due to the involved computational cost: at a resolution allowing identification of individual coral colonies, the largest \new{high-resolution} 3D reconstructions cover 60 m in length, while taking days of computation time \citep{reefscapegenomics}.
	Similarly, systems for benthic cover classification are commonly restricted to photo quadrats or orthomosaics as opposed to general images of reef scenes, and are often only trained on datasets with sparse pixel annotations \citep{coralnet, coralnetengine, yuval2021repeatable}. 
	The state-of-the-art is far from general-purpose semantic segmentation systems for coral reef scenes that work reliably across reef scenarios and conditions. \new{Furthermore, it can be challenging to transfer results from benthic cover estimation into 3D reconstructions made with photogrammetry \citep{reverse}.} These limitations, \new{and the manual labor to overcome them}, have impeded computer vision tools from being applied under water at the same scale as in terrestrial settings.
	\vspace{3pt}

	\begin{figure*}[t!]
		\raggedleft 
		\includegraphics[width=\linewidth, height=150px]{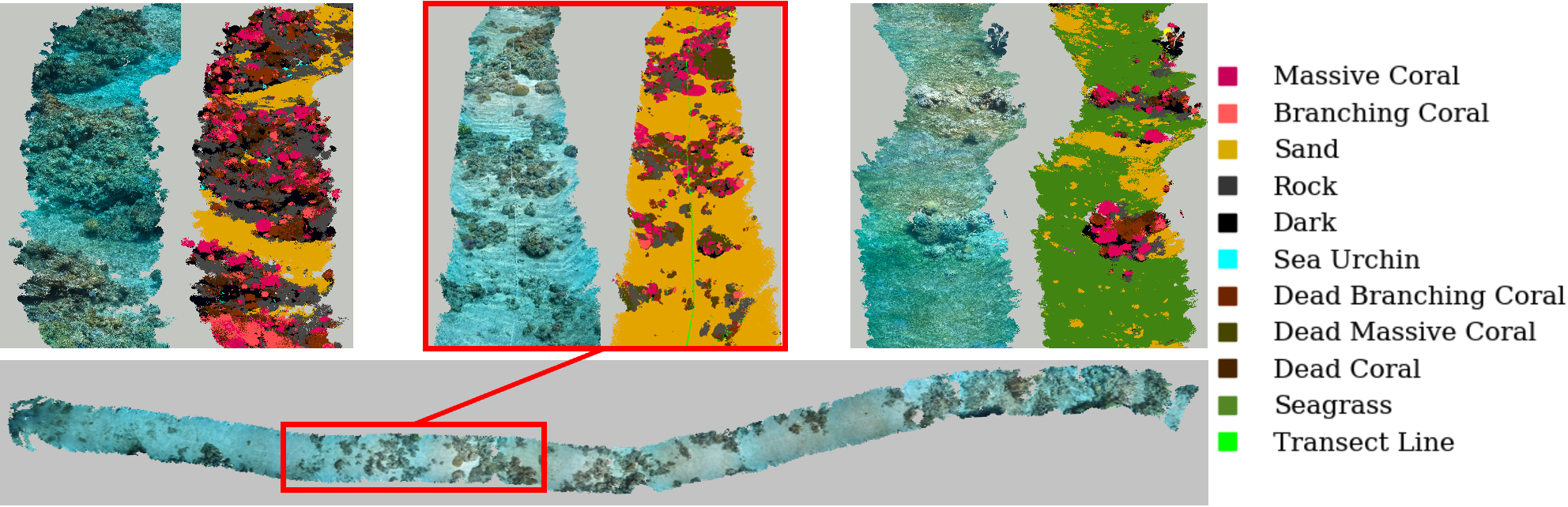}
		\caption{Example excerpts of 3D point clouds of different reef scenarios in their original RGB color, next to the points colorized by their predicted benthic class (top). A 100 m transect (bottom) can be covered by a diver in less than five minutes: the length of the created point clouds is limited only by the diver's air availability and the camera's battery capacity.}\label{fig:scaleup} 
	\end{figure*}

	In this paper, we present an approach for 3D mapping and simultaneous semantic segmentation of coral reef areas that is significantly more scalable than existing approaches. The underlying paradigm of learning-based SfM photogrammetry uses deep neural networks that learn to adapt to challenging environments for computer vision algorithms, such as those posed under water. In particular, we show that it is possible to create 3D maps of large areas of reef at resolution of individual coral colonies using a single affordable consumer-grade camera filming while being moved around the scene ('ego-motion' video). Our method requires no expensive computing infrastructure: on a computer with a single Graphics Processing Unit (GPU), the semantic segmentation and 3D reconstruction can be obtained in real video time. From a single SCUBA dive, it is possible to obtain a 3D point cloud of more than 1km length at the resolution of individual coral colonies, as shown in Figure \ref{fig:scaleup}.

	The proposed 3D mapping approach enjoys a powerful synergy with image-based semantic segmentation systems, which assign each pixel in an image to a specific semantic class. The semantic information can be directly transferred to the 3D models which in turn enables automated computation of \new{ecological measures} of interest, such as the area covered by each benthic class.
	
	We exemplify our approach, which we name DeepReefMap, on reef areas in the northern Gulf of Aqaba, publishing a large-scale dataset of ego-motion videos taken by divers in the area. Furthermore, we publish a dataset of video frames of reef scenes annotated for pixel-wise semantic segmentation of benthic classes. We train neural networks to reconstruct the 3D geometry from the videos and for semantic segmentation of 20 benthic classes. \new{We find that the accuracy of the estimated 3D geometry from our method is competitive with a state-of-the-art conventional SfM pipeline, whereas our approach is more robust and two orders of magnitude faster.
		The semantic segmentation system is} evaluated on three scenes that were not seen during training, where 84.1\% of pixels are correctly classified. Our implementation is open source, alleviating the reliance on proprietary SfM software.

	\section*{Materials and Methods} 
	
\begin{figure*}[t!]
		\includegraphics[width=\linewidth, right, height=153px]{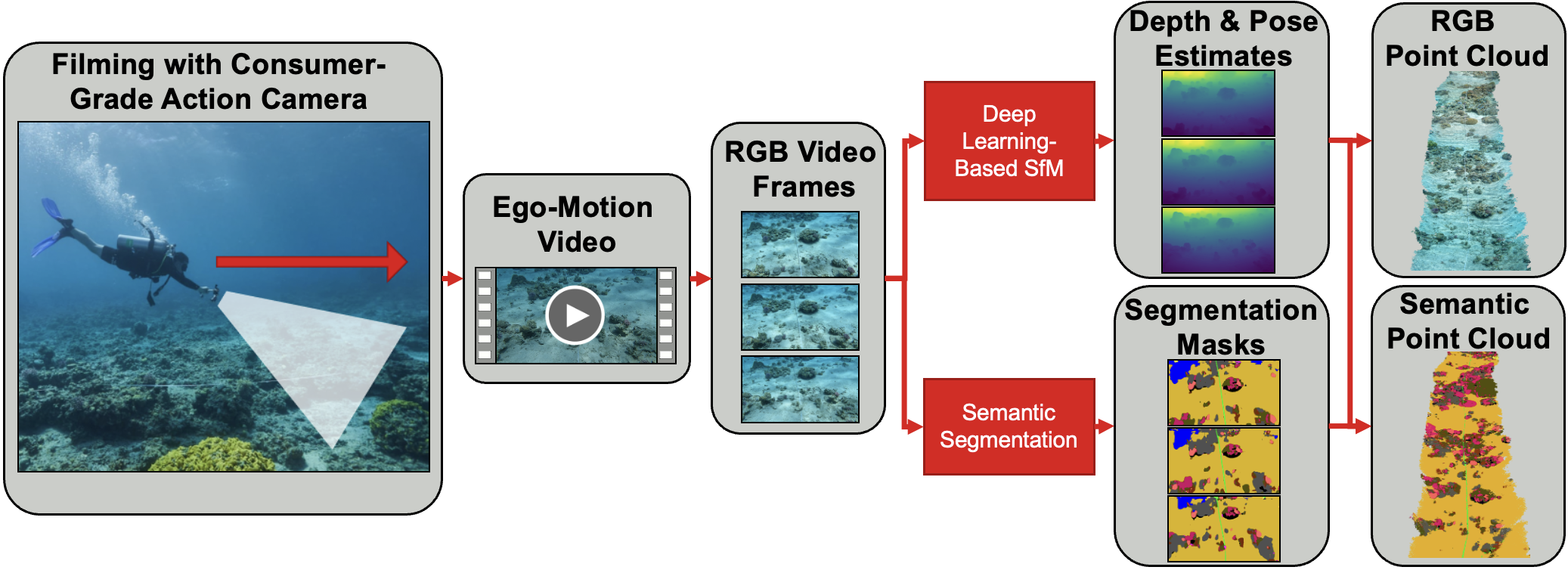}
		\caption{An overview of the main method (DeepReefMap): a diver swims over a reef while taking video with a consumer-grade action camera. On the uncurated frames of this video, the learning-based SfM system is used to reconstruct the 3D geometry of the reef in real time. A semantic segmentation system is used to identify benthic classes, which can directly be transferred into the point cloud.}
		\label{fig:pipeline}
	\end{figure*}

	DeepReefMap creates 3D point clouds from uncurated ego-motion videos by using deep learning-based SfM and leveraging its synergy with semantic segmentation of benthic classes. An overview is shown in Figure \ref{fig:pipeline}. The remainder of this section describes in detail the data collection process, the learning-based SfM system and the semantic segmentation system.
	
	\subsection*{Collection of Ego-Motion Videos}
	
	The collection of data from which 3D maps are created is particularly straightforward: a diver swims forward while being between 1 and 5 meters above a reef area, filming with a consumer-grade action camera as illustrated in Figure \ref{fig:pipeline} (left). \new{From a technical side, our method is agnostic to the video camera used. Consumer-grade action cameras were chosen because they are the cheapest options.} The distance covered is only limited by logistical constrains such as the diver's air reserves, the battery and memory capacity of the camera, the length of the reef, or the distance to the diving boat. Within a single SCUBA dive, a video covering more than 1 km of distance can routinely be taken.
	
	For this work, the dataset of ego-motion underwater video used was collected by divers from six reef sites in Eilat (Israel) and Aqaba (Jordan) between July 22, 2022 and Aug 25, 2022, on an expedition of the Transnational Red Sea Center \footnote{http://trsc.org}. 
	The videos cover a diverse set of scenes: from reefs with high structural complexity, over patchy reefs in seagrass meadows, to sites located close to human settlements, which are heavily exposed to human activity and the resulting pollution. \new{The videos were collected in reefs ranging from 3 to 10 m in depth, with no particular constraint on the swim speed. Three 100 m transects were filmed for benchmarking purposes, while the remaining videos are taken without any reference objects placed on the substrate in a free-roaming fashion.}
	
	The videos were captured with GoPro Hero 10 cameras in the linear field-of-view mode at 1080p resolution and 30 frames per second with the stabilization setting on `smooth'. In order to increase the amount of video data captured during each dive, three cameras were attached side-by-side onto a rigid pole, with about 1m of space in between each camera. In total, there are 19 hours and 49 minutes of ego-motion video. 
	
	The cameras were attached using simple handlebar mounts, and re-attached by hand before every dive without highly precise angle measurement, leading to a slight variation in camera angles between dives. Up to the precision of eye-balling the angles, the roll and yaw angles of the cameras were fixed at $0^{\circ}$. The pitch angle was deliberately varied between approximately $-5^{\circ}$ and $-40^{\circ}$ between dives in order to include diverse settings into the training dataset.

	\subsection*{Deep Learning-Based SfM}
	
	The 3D mapping in our approach is realized using deep learning-based SfM \citep{sfmlearner, scaleconsistent}, in which neural networks are trained from example videos to specialize SfM photogrammetry to the challenging conditions to which it is exposed to in a reef environment. As opposed to conventional SfM, where strong assumptions on the color consistency of image features are made, leading to brittle reconstruction behavior, learning-based SfM can adapt to the challenges of underwater scenes.
	Learning-based SfM leverages the geometric relationship between overlapping images to formulate an unsupervised learning objective, meaning that it suffices to use a large dataset of unannotated video for training such a system, without ground-truth camera position or depth data.

\begin{figure}[t!]
		\includegraphics[width=0.99\linewidth, right]{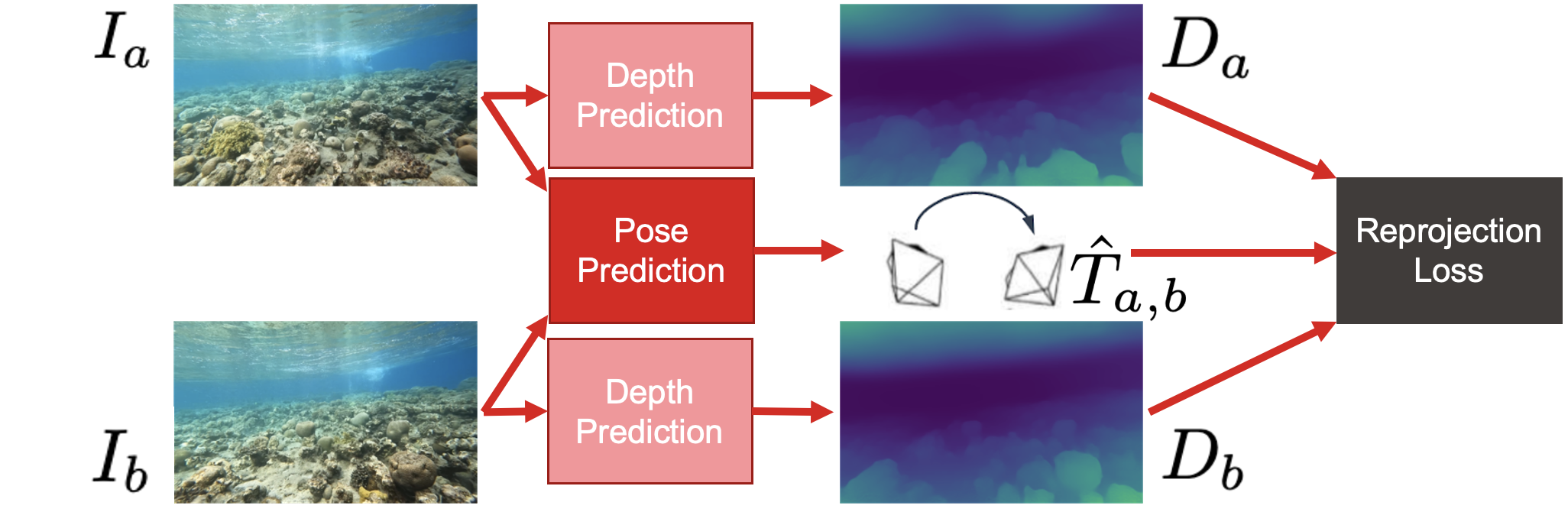}
		\caption{Schematic overview of a deep learning-based SfM setup. A neural network takes two \new{overlapping} images $I_a$ and $I_b$ and estimates their respective depths as well as the camera pose transformation between them. These estimates are used to project one image onto the other and compute a regularized photometric reprojection error, which acts as an unsupervised learning signal to train the involved neural networks.} 
		\label{fig:learningbasedsfm}
	\end{figure}

	The underlying principle of learning-based SfM is to estimate the rigid transformation between pairs of images, formulating a differentiable loss function that provides a learning signal for a neural network. A schematic overview of a learning-based SfM system is shown in Figure \ref{fig:learningbasedsfm}. Consider two \new{overlapping} images $I_a$ and $I_b$ taken with the same camera at different locations and orientations. Let $\hat{T}_{a,b}$ denote an estimate of the 6D camera pose transform (the 3D translation and 3D rotation) between the camera poses of the two images. Using an estimate of the depth $\hat{D}_a$ of an image, i.e. the distance between the camera and each pixel, 
	and choosing an appropriate camera model, one can take an image and its depth estimate to reproject $I_a$ to the camera position of $I_b$, obtaining a reprojected image $\hat{I}_b$:
	
	\begin{equation}
		\hat{I}_b = \text{Reproject}(I_a, \hat{D}_a, \hat{T}_{a, b})
	\end{equation}

	The reprojected image can then be compared to the original image, using a differentiable photometric loss function between the images in RGB space, denoted $\mathcal{L}_{P}$. This reprojection loss is calculated in both directions, by using $\hat{D}_b$ and the inverse camera pose transform $\hat{T}_{b,a} = \hat{T}_{a,b}^{-1}$. In a similar fashion, the estimated depths can be reprojected via the estimated pose transform and a regularization term $\mathcal{R}_G$ that enforces their geometric consistency is added to the loss function \citep{scaleconsistent}.
	A smoothness regularization term $\mathcal{R}_S$ on the estimated depths is added to deal with low-texture regions \citep{sfmlearner}. In state-of-the-art systems, a similar image-pair unsupervised formulation of optical flow is included in the training procedure, which we omit here for the sake of simplicity. The estimates of the depth and camera pose transform are predicted from the pair of images by a neural network $f$, which has a set of learnable parameters $\theta$:
	
	\begin{equation}
		\hat{D}_a, \hat{D}_b, \hat{T}_{a,b} = f_\theta(I_a, I_b)
	\end{equation}

	Using the photometric loss and the regularization terms, the parameters of the neural network are then trained by gradient-based optimization to minimize the total expected loss over the data, which is empirically realized by training on the dataset of available pairs or overlapping images:
	\begin{equation}
		\begin{split}
			\min_\theta \mathbb{E}_{I_a, I_b} \Big[ &\mathcal{L}_{P}(I_a, \hat{I}_a) + \mathcal{R}_{{S}}(\hat{I}_a,\hat{D}_a) + 
			\mathcal{L}_{P}(I_b, \hat{I}_b) +  \mathcal{R}_{{S}}(\hat{I}_b,\hat{D}_b) + \mathcal{R}_{{G}}(\hat{D}_a, \hat{D}_b, \hat{T}_{a, b}) \Big]
		\end{split}
	\end{equation}
	Training a neural network on a sufficiently large and diverse dataset of videos enables accurate depth and pose estimation for the scenarios depicted in the training videos: the SfM system learns from examples to be robust against the challenging conditions posed by coral reef settings.

	For training our learning-based SfM system, video frames were extracted from the dataset of ego-motion videos at a resolution of $608\times352$ px with 20 frames per second, leading to a total of 1.4 million video frames for training the system. Our implementation is based on the SC-SfMLearner \citep{scaleconsistent}, with the neural network architecture chosen to be a U-Net \citep{unet} with a ResNet34 \citep{resnet} backbone, which is selected due to its large receptive field size but fast inference speed. This neural network is trained for 1 million steps of batch size 5 using the Adam optimizer \citep{adam} with a learning rate of 0.0001. More implementation \& training details can be found in the appendix.

\begin{figure}[t!]
		\centering
		\includegraphics[width=\linewidth]{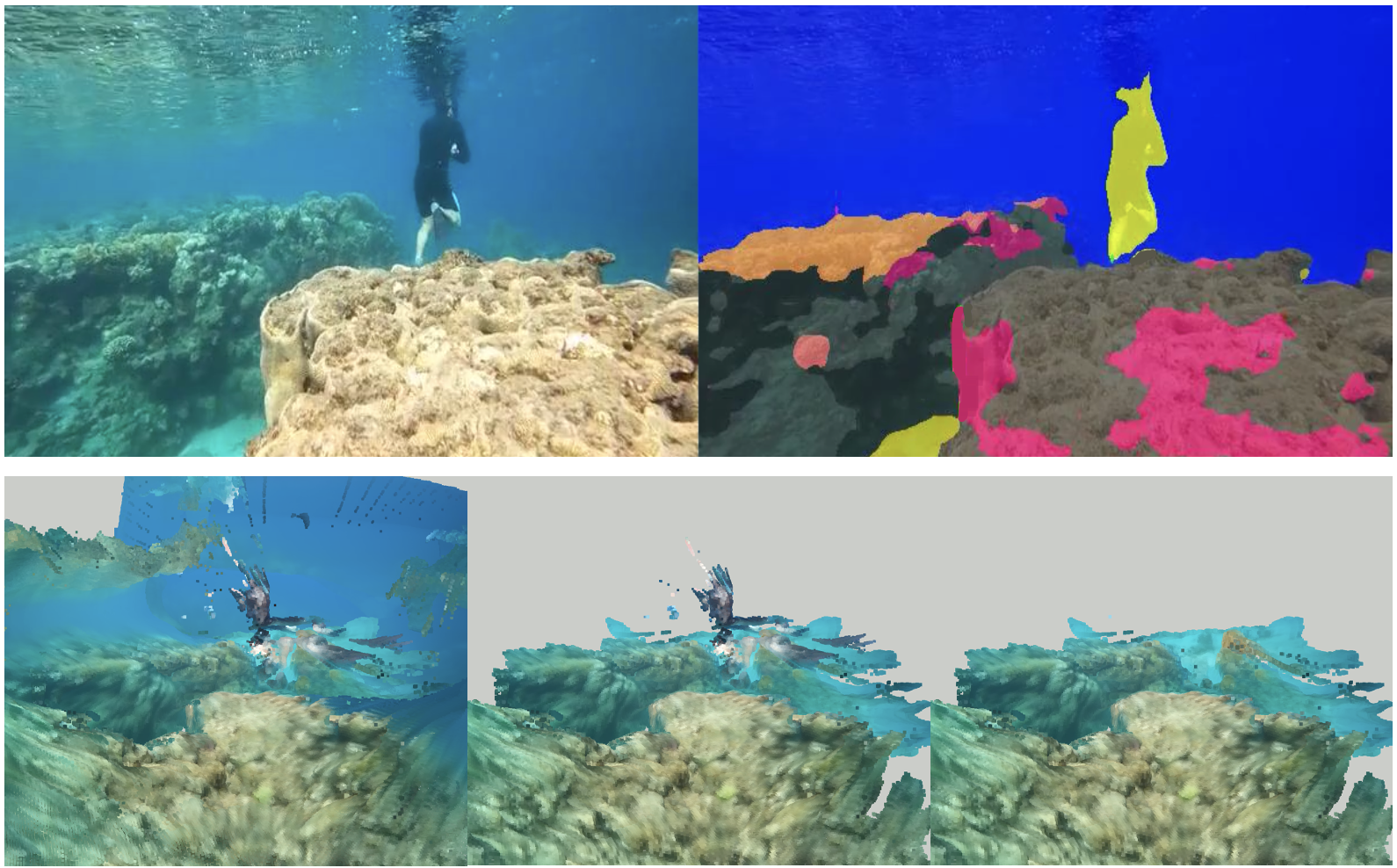}

		\caption{\new{Using a trained semantic segmentation system, unwanted classes such as the background (dark blue) and human (in yellow) can be masked out in the point cloud creation process (top). This can alleviate artifacts from these classes, as illustrated by this example from the Japanese Garden Site in Eilat, showing the point cloud made with all classes (bottom left), the background removed (bottom center), and both unwanted classes removed (bottom right).}}
		\label{fig:beforeafter}
	\end{figure}

	Using a trained learning-based SfM system to create a 3D point cloud can be realized by iterating through all the frames of a video, selecting some or all pixels of each frame, projecting them out into 3D space using the estimated depth and the camera intrinsics, and finally updating the camera position using the estimated camera pose transform. This can be done essentially at the speed of the forward pass of the trained neural networks: a U-Net with a ResNet-34 backbone can create a 3D point cloud from a video at 18 frames per second - essentially real video time. \new{For better visual quality, the frames are added to the point cloud via integration though a truncated signed distance field \citep{tsdf}, a common technique in dense monocular 3D mapping. Naively integrating} frames of an underwater video with \new{this iterative technique} will however lead to strong undesirable artifacts from the water surface, the background water column, and dynamic objects, such as fish and divers, as displayed in Figure \ref{fig:beforeafter}. As a remedy, semantic segmentation can be employed.

	\subsection*{Semantic Segmentation}
	
	\begin{figure}[hb!]
		\centering
		\includegraphics[width=0.96\linewidth]{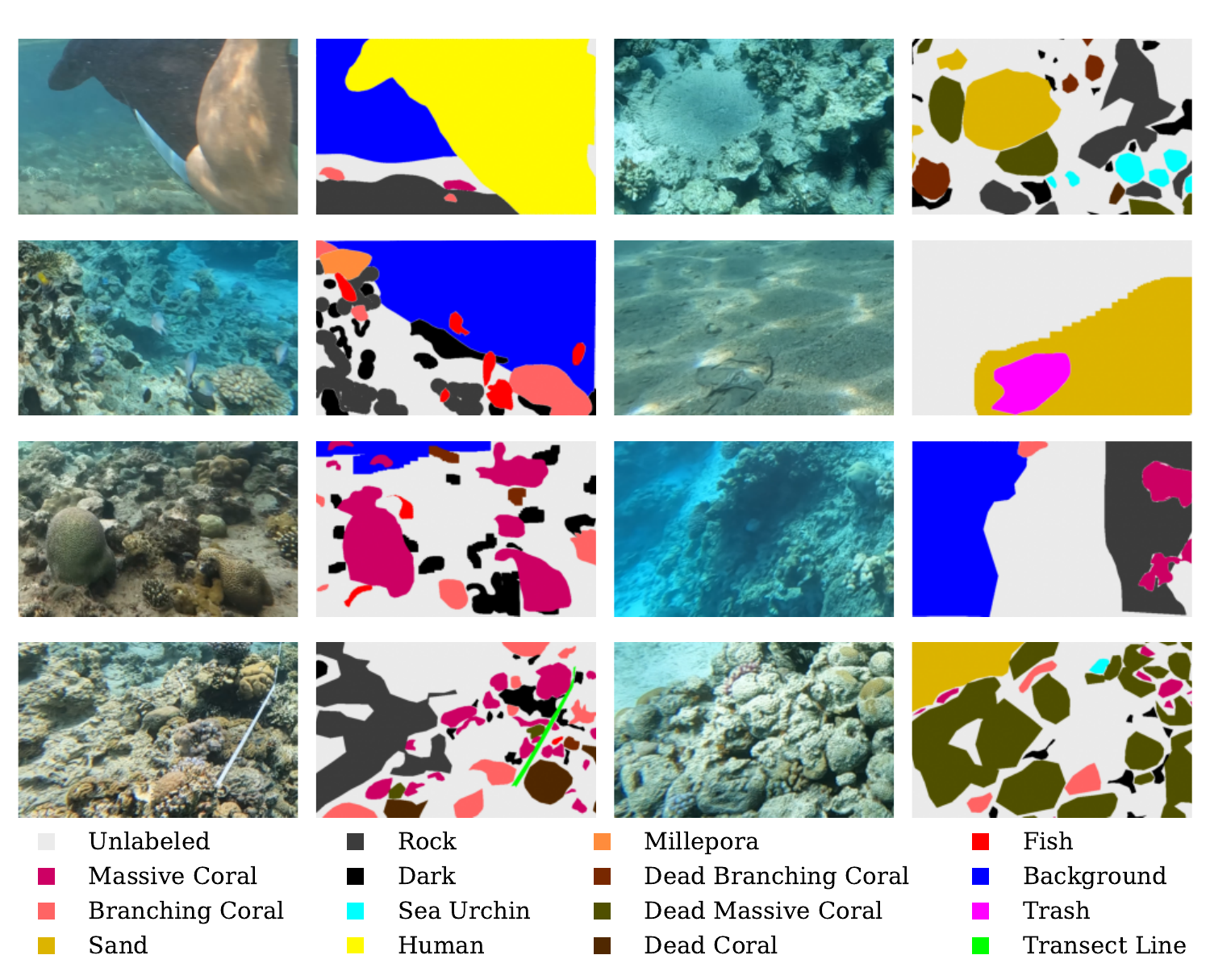}
		\caption{Example video frame patches of size $800\times500$ px and their annotations. In many video frames with annotations, portions of the image remain unlabeled.}
		\label{fig:annotatedpatches}
	\end{figure}
	
	There is a strong synergy between learning-based SfM systems and image-based semantic segmentation: information about the benthic composition can be transferred to the point cloud and unwanted classes that lead to artifacts in the point clouds can be removed. This Section describes our semantic segmentation system and dataset, as well as its interaction with the learning-based SfM system.
	
	We created a dataset of video frames that were annotated with 20 benthic classes of interest. In total, 1997 patches of size $800\times500$ px were annotated for semantic segmentation with over 10000 polygons using the AIDE annotation software 2.0 \citep{aide}, which extracts patches from the full $1920\times1080$ px video frames. Example patches and their labels are shown in Figure \ref{fig:annotatedpatches}, showing how polygons are drawn around objects and some areas contain no label. Details on the label distribution from the 20 classes are provided in the appendix. Our dataset is the first dataset containing polygon-level benthic segmentation labels of `general purpose' reef scenes when compared to existing datasets in the coral reef domain, which have either focused on specific subdomains, such as photo quadrats \citep{coralnet} or orthomosaics \citep{yuval2021repeatable}, or only provided sparse point labels instead of segmentation labels \citep{coralnet}.
	
	The dataset of video frame patches contains patches from the three evaluation transects, where the patches were chosen to maximize spatial coverage and minimize spatial overlap between frames for each transect. The dataset also includes patches chosen from various scenes, which were selected manually to get a good coverage of the label classes and of the diversity of the scenes from which video was captured but have no spatial overlap with the evaluation transects.

	\paragraph{Inherent Ambiguity} While some classes, such as colorful live corals or fish, have distinct features that can be labeled consistently, other classes have a strong inherent ambiguity. In particular, it can be extremely hard even for expert human annotators to draw the line between rock, macroalgae-covered substrate, dead corals, and rubble. Furthermore, while some live coral is obviously alive and some dead coral is obviously dead, there are cases in which it is difficult to assign a health label for damaged corals. While maximal accuracy would necessitate an underwater side-by-side comparison of the pigmentation of the coral with a printed color chart and a lookup in benchmark pigmentation charts of the exact species at hand, the annotation was instead performed with best judgment of the annotators in such ambiguous cases. Lastly, many animals, such as fish and sea urchins, take shelter in holes. Due to the limited lighting, often only parts of the animal are discernible, and it is up to the annotator's judgment where the animal ends and the class `dark' begins.\\

	\noindent\textbf{Training the Neural Network}\hspace{3pt}
	On the dataset of annotated video frames, we train a semantic segmentation neural network. In particular, we use the U-Net \citep{unet} architecture with a ResNeXt50-32x4d \citep{resnext} backbone. The input and output resolutions are chosen to be $416\times416$ px. The backbone is initialized with weights pre-trained on the large-scale dataset ImageNet \citep{imagenet}. 
	The U-Net is trained to minimize the pixel-wise cross-entropy between the classes using the Adam \citep{adam} optimizer with a learning rate of 0.0001 and a batch size of 32. Unlabeled pixels are omitted from the learning objective. Data augmentations are used to artificially increase the size of the dataset: images are randomly flipped horizontally, and randomly resized crops with scales between 0.5 to 1.4, and aspect ratio modifications between 0.7 and 1.4 are chosen. 
	When using the trained semantic segmentation network to calculate the benthic classes of images, the full $1920\times1080$ px image is divided into overlapping patches of size $800\times500$ px, which are resized to $416\times416$ px before being fed into the neural network, obtaining segmentations of the same size. \new{The segmentation predictions} are resized back to 
	$800\times500$ px and re-stitched together \new{(in image regions from overlapping predicted patches, the softmax class probabilities are averaged)} to yield a semantic segmentation in the original video resolution. \\
	
	\noindent\textbf{Use within learning-based SfM}\hspace{3pt}
	The creation of 3D point clouds from images using a learning-based SfM system iterates through the video frames, projecting image pixels with their depth into 3D space. With a trained semantic segmentation system, unwanted classes such as divers, fish, and background, can simply be excluded during this procedure by masking them out at the image level, as shown in Figure \ref{fig:beforeafter}. For the remaining benthic classes of interest, the benthic class of the pixel is attached to the respective point in the 3D point cloud, facilitating automation of downstream ecological analysis. 
	
	\subsection*{Evaluation}
	\new{\paragraph{Spatial Accuracy} To evaluate the spatial accuracy of the 3D point clouds produced by DeepReefMap, another $38$ m long video transect was collected on Aug 4, 2023 in the Japanese Garden in Eilat with placed ground markers. Divers measured the ground-truth distance between targets. We evaluate the accuracy of our method against the ground-truth and compare against the research standard COLMAP \citep{colmap} and the industry software Aigsoft Metashape \citep{agisoft}. In particular, ten markers were placed and twelve distances between 70 cm and 370 cm were measured by hand. Following this, an ego-motion video of this transect was taken with the default setup of our method, leading to a video of 2 minutes and 24 seconds. Then, top-down pictures of the same transect were taken with the GoPro camera in linear photo mode (at $5568 \times 4176$ px resolution), following the protocol from \cite{raoult2016gopros}. The top-down images acquisition took 6 minutes. These photos are taken in a way that excludes background haze, other divers, or large fish. DeepReefMap takes images of size $608\times352$ px as input, so the top-down images are scaled and cropped accordingly (maintaining the aspect ratio) to be used as an input.}
	
	\new{We provide quantitative results as the mean absolute relative error from the ground truth distances:}
	
	$$\text{Mean Absolute Relative Error} = \frac{1}{|D|}\sum_{i, j \in D} \big| d_{ij} - \hat{d}_{ij}\big|\big/d_{ij}.$$
	
	\new{Here $D$ is the set of pairwise ground truth distances, $d_{ij}$ is the ground truth distance between target $i$ and target $j$, and $\hat{d}_{ij}$ is the measured distance using the ruler tool in Metashape. Before this, the coordinates of the point cloud from DeepReefMap are multiplied by a scalar factor of $\text{mean}(\hat{d})/\text{mean}(d)$ in order to scale the point clouds to comparable size.}

	\paragraph{Semantic Segmentation} To evaluate the semantic segmentation system of our method, we evaluate on three 100 m line transects: two transects from different sites in the King Abdullah Reef in Jordan, and one from the Japanese Garden reef in Israel. We start by evaluating the semantic segmentation system both qualitatively and quantitatively, followed by a qualitative analysis of the 3D reconstructions and the respective automatic downstream ecological analysis.
	
	We separate the annotated video frames into a train and test dataset for each of the three evaluation transects, with the test set being formed by all annotated frames of the transect. \new{This way, the transect on which we evaluate is unseen during training and there is definitely no overlap between the train and test images, thus giving an estimate to the system's generalization performance on new data.}

	\paragraph{Ortho-Projection} Ortho-projected 2D maps from the 3D point clouds are created as follows:
	first, a 2D occupancy grid on which the gravity vector (obtained from the camera's inertial measurement unit) is the normal vector is chosen with a suitable grid cell size. In each 2D cell, the 30\% of 3D points in the cell with the highest z-value (the lowest depth) are selected. To increase robustness against noisy points, the benthic class for the grid cell is chosen by \new{(hard)} majority voting of points present in the grid cell, whereas the RGB value and z-value for the grid cell are obtained through averaging the points. The true scale of the objects in the point cloud and 2D map can be obtained by scaling with objects of known reference size, for example a transect line.

	\vspace{-5pt}
	\section*{Results}
	\vspace{-5pt}

	\new{\textbf{Spatial accuracy}} We present large-scale semantic 3D maps of coral reefs from six sites in the northern Gulf of Aqaba, created from video taken with one consumer-grade action camera. The output 3D point clouds have the semantic segmentation transferred directly from the video frame pixels to the 3D points, which allows automatic downstream benthic cover estimation. 
	Each 3D map is produced directly from a single video in real time, \new{speeding up the analysis of video transects} by orders of magnitudes compared to previous methods.

	\begin{table*}[b]
		\centering
		\caption{\new{Spatial accuracy with regards to ground-truth markers placed in the Japanese Garden video of Aug. 4, 2023. The cells marked with * denote point clouds which are locally coherent and complete, but not globally coherent.}}
		\resizebox{\textwidth}{!}{%
			\begin{tabular}{l|ccc|ccc}
				\toprule
				Method & \multicolumn{3}{c|}{Mean Absolute Relative Error} & \multicolumn{3}{c}{Processing Time Taken (seconds)}\\
				\midrule
				& Ego-Motion & \makecell{Top-Down\\ (Low-Res)} & \makecell{Top-Down\\ (High-Res)} & Ego-Motion & \makecell{Top-Down\\ (Low-Res)} & \makecell{Top-Down\\ (High-Res)}\\
				\midrule
				DeepReefMap (ours) & 7.84\% & 9.84\%$^*$& - & \textbf{320} & \textbf{100} &
				-\\
				Agisoft Metashape & Fail & Fail & \textbf{1.85\%} & 9 870 & 1 800 & 21 850 \\  
				COLMAP & Fail & Fail & 6.58\%$^*$ & 75 660 & 1 550 & 47 520 \\  
				\bottomrule
		\end{tabular}}
		\label{table:spatial}
	\end{table*}

	\new{A quantitative evaluation of the spatial accuracy is shown in Table \ref{table:spatial}. DeepReefMap is the only method that produces a point cloud encompassing all frames in the low resolution setting. The error of DeepReefMap in the ego-motion video setting is lower than in the top-down setting and is comparable to the error of the performance of COLMAP in the high-resolution top-down setting, which, however, took more than $100$ times longer to compute. Note that both Metashape and COLMAP fail in reconstructing the point cloud in the low resolution and ego-motion settings despite the high overlap between images, and the area covered being relatively small ($38$ m).}

\begin{figure}[t]
		\centering
		\includegraphics[width=0.94\linewidth]{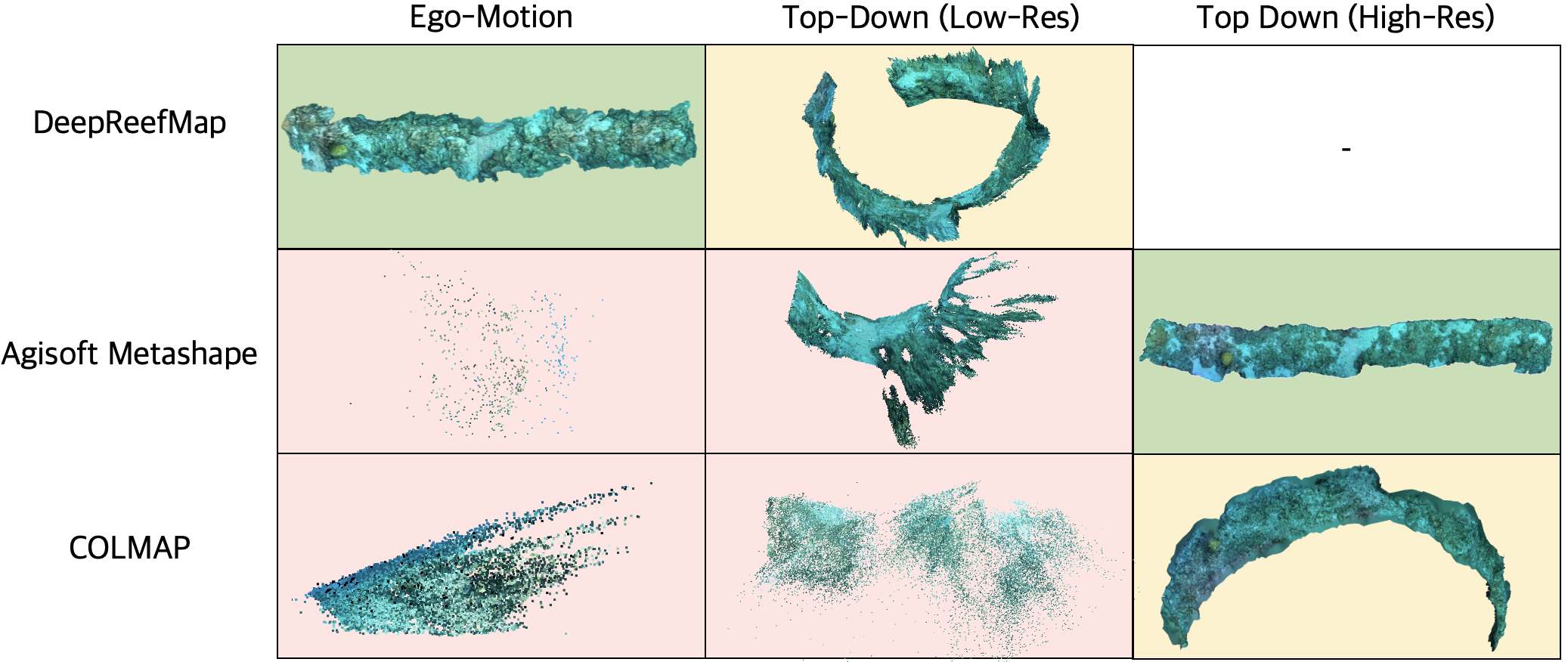}
		\vspace{-2pt}
		\caption{\new{Comparison of 3D Reconstructions using different input imagery. The top-down images are either of high resolution (last column, $5568\times4176$ px), or are downscaled to the lower input resolution of DeepReefMap (center column, $608\times352$ px). In general, only DeepReefMap and Metashape produced globally coherent point clouds. When compared low resolution imagery, only DeepReefMap on ego-motion video yields satisfactory results. }}
		\label{fig:spatial_comparison}
	\end{figure}

	\new{This is further highlighted in Figure \ref{fig:spatial_comparison}, where the resulting point clouds are visualized for all methods. DeepReefMap and Metashape in the high resolution settings are the only ones providing both locally and globally coherent reconstructions. Metashape with the high-resolution images produces the best final spatial accuracy (see Table~\ref{table:spatial}) at the cost of needing high resolution images and very long computation times, while DeepReefMap has much faster data acquisition and reconstruction time, and allows directly transfering semantic segmentation from the video frames to the 3D point cloud. COLMAP in the high-resolution setting and DeepReefMap in the top-down low-resolution setting produce locally coherent point clouds that are not globally coherent, and are twisting inside themselves. The top-down images have an overlap of around 80-90\%, making the pose between frames much bigger than for subsequent video frames. When the camera intrinsics do not account for the diffraction caused by even linear cameras under water (if no dome port is used), the re-projection is minimized by overestimating the rotation, leading to the characteristic rounded shape. For COLMAP, this happens despite the camera intrinsics to be set to a radial model. DeepReefMap for now assumes a linear camera model, and the gravity vector is not saved in GoPro images (which could be used to alleviate the twisting), unlike in videos. }

	\begin{figure}[b!]
		\centering
		\includegraphics[height=81px]{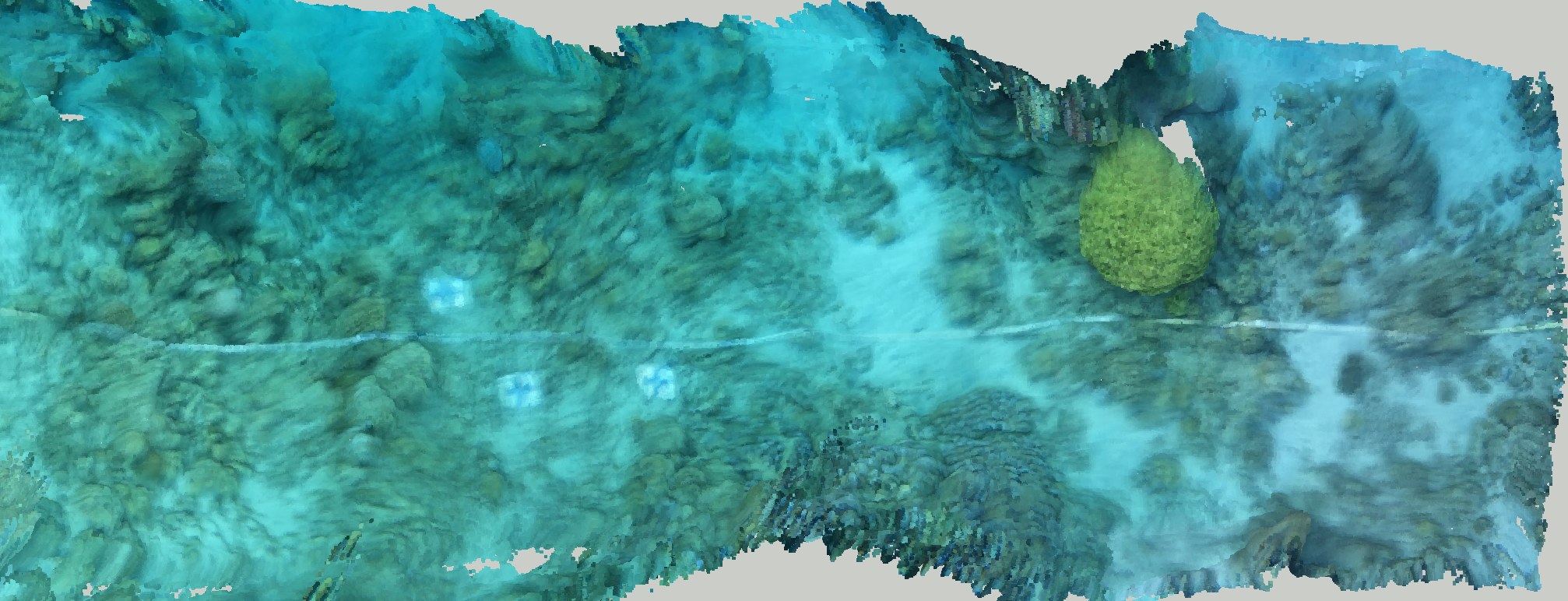}
		\includegraphics[height=81px]{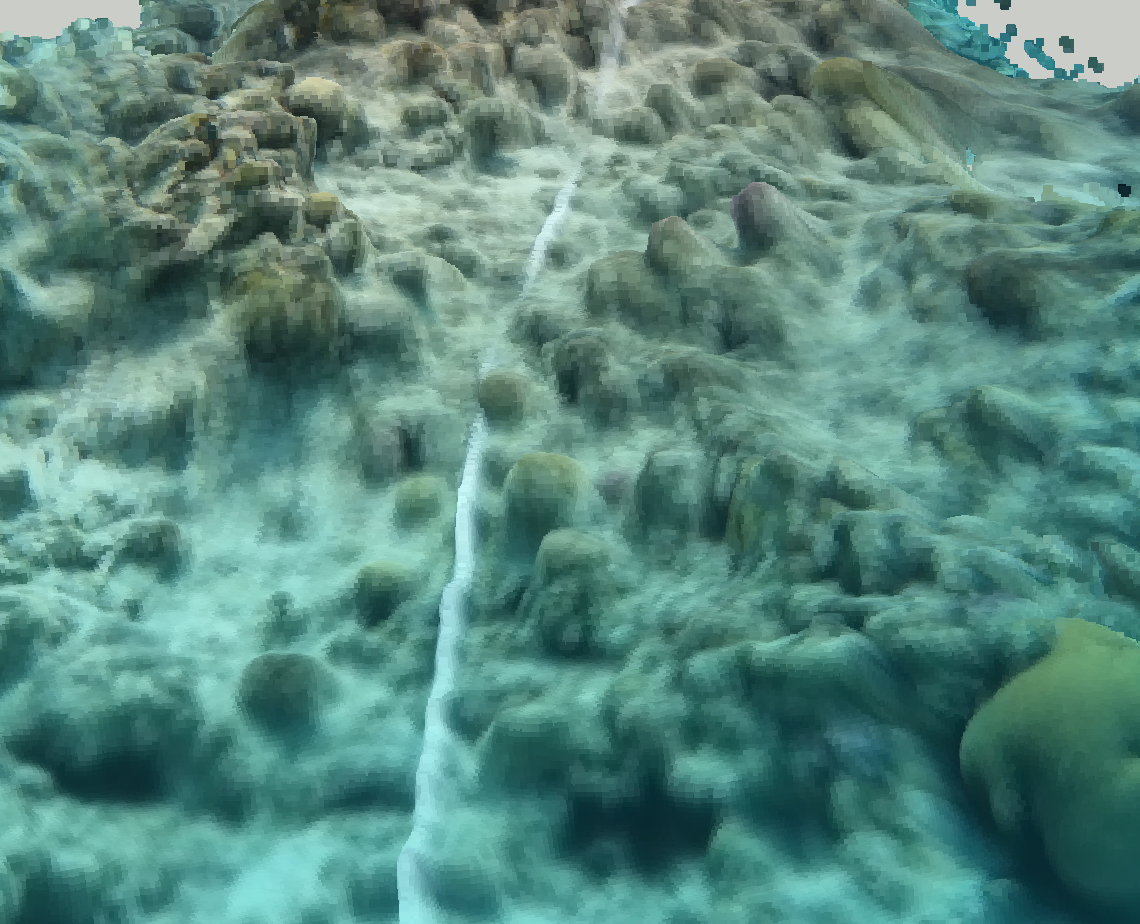}
		\includegraphics[height=81px]{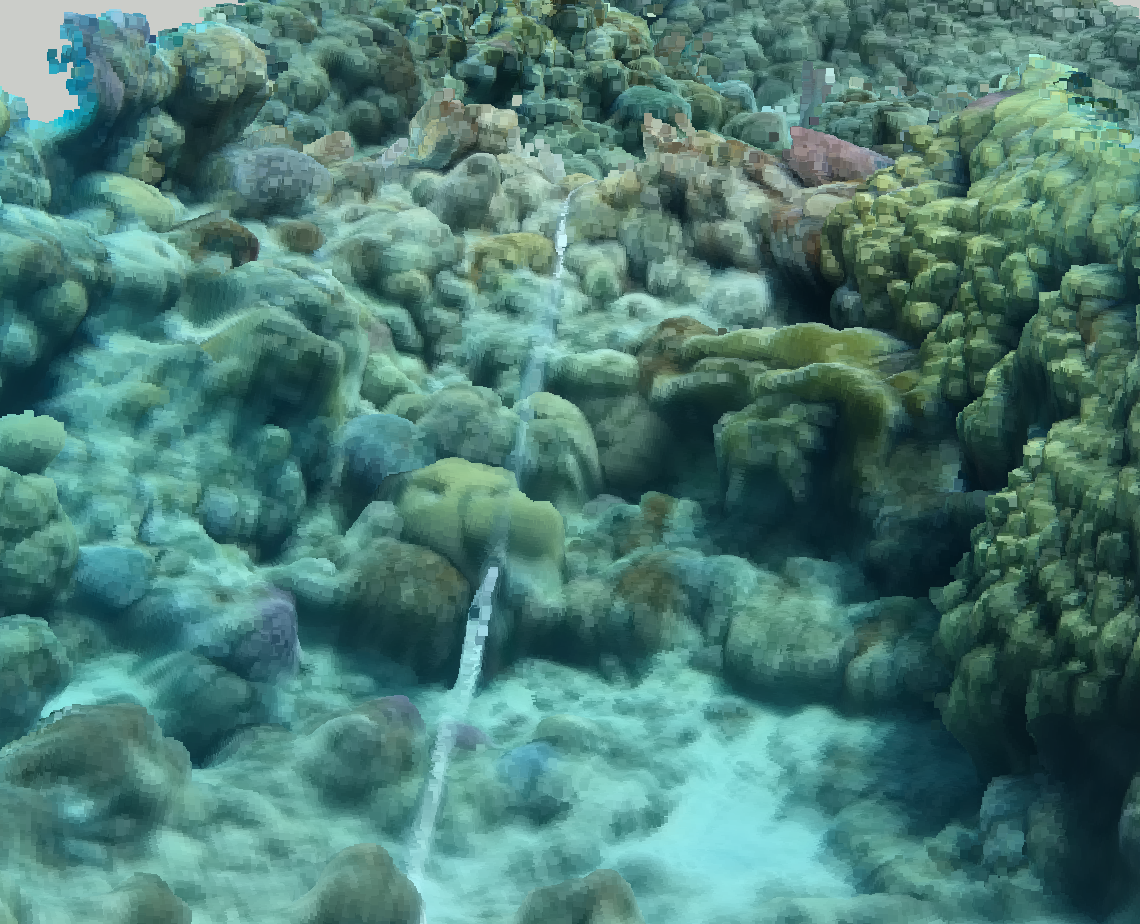}
		\includegraphics[height=88px]{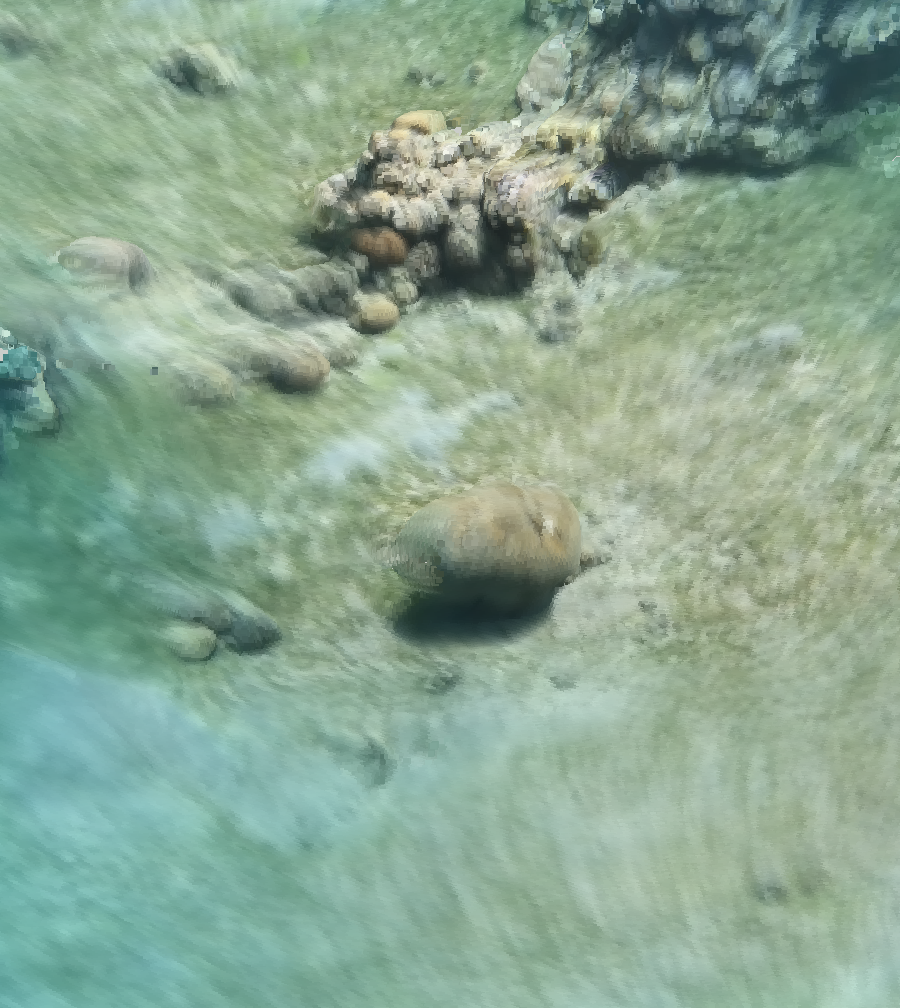}
		\includegraphics[height=88px]{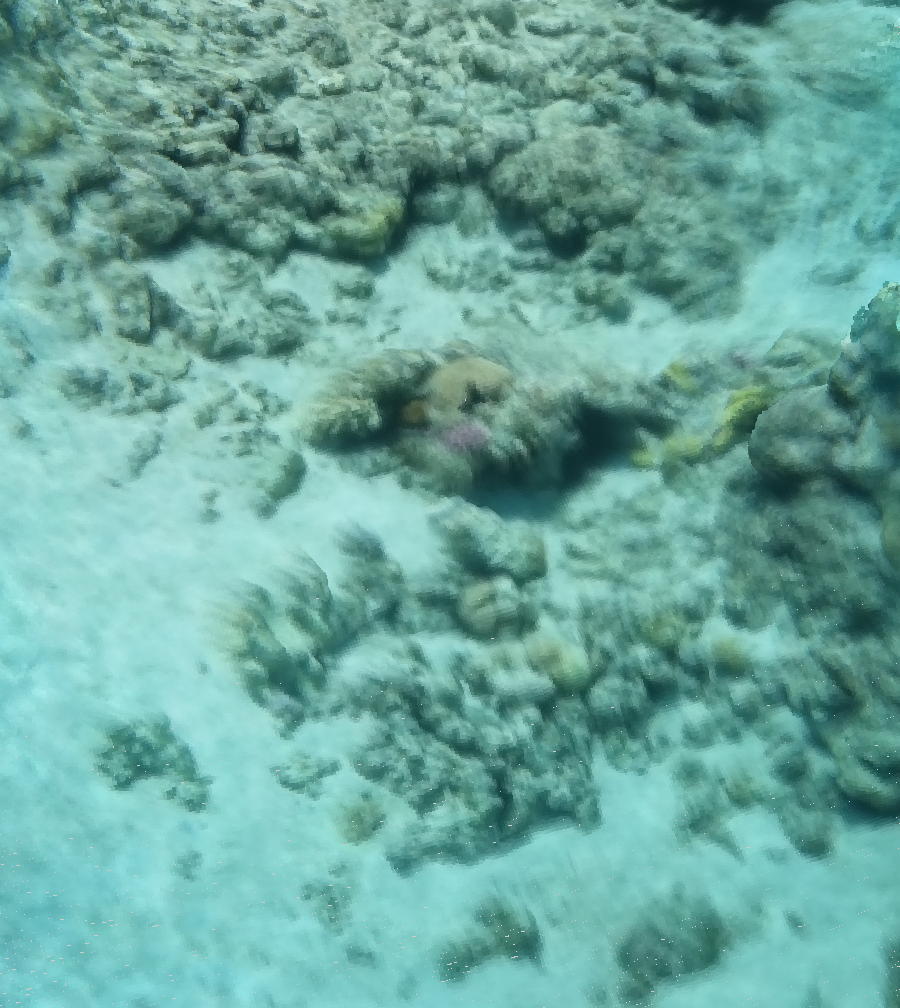}
		\includegraphics[height=88px]{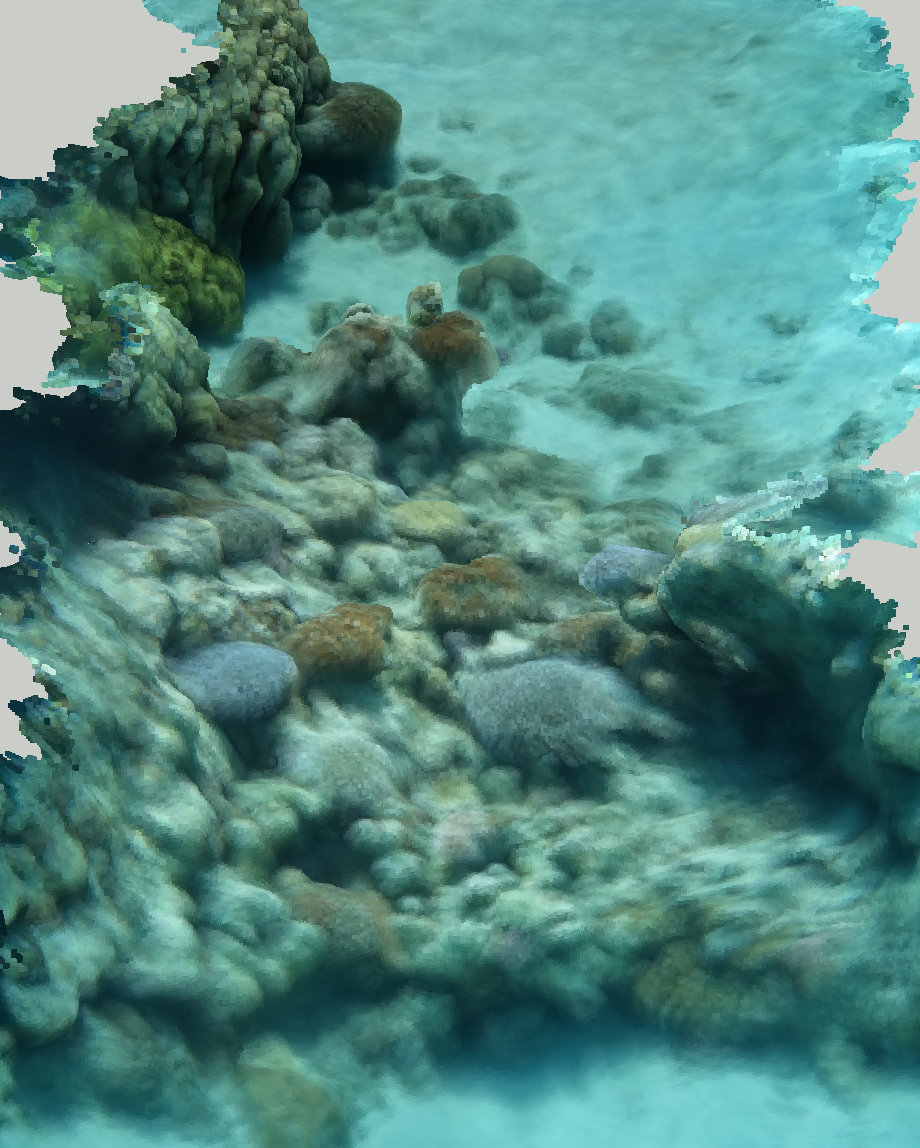}
		\includegraphics[height=88px]{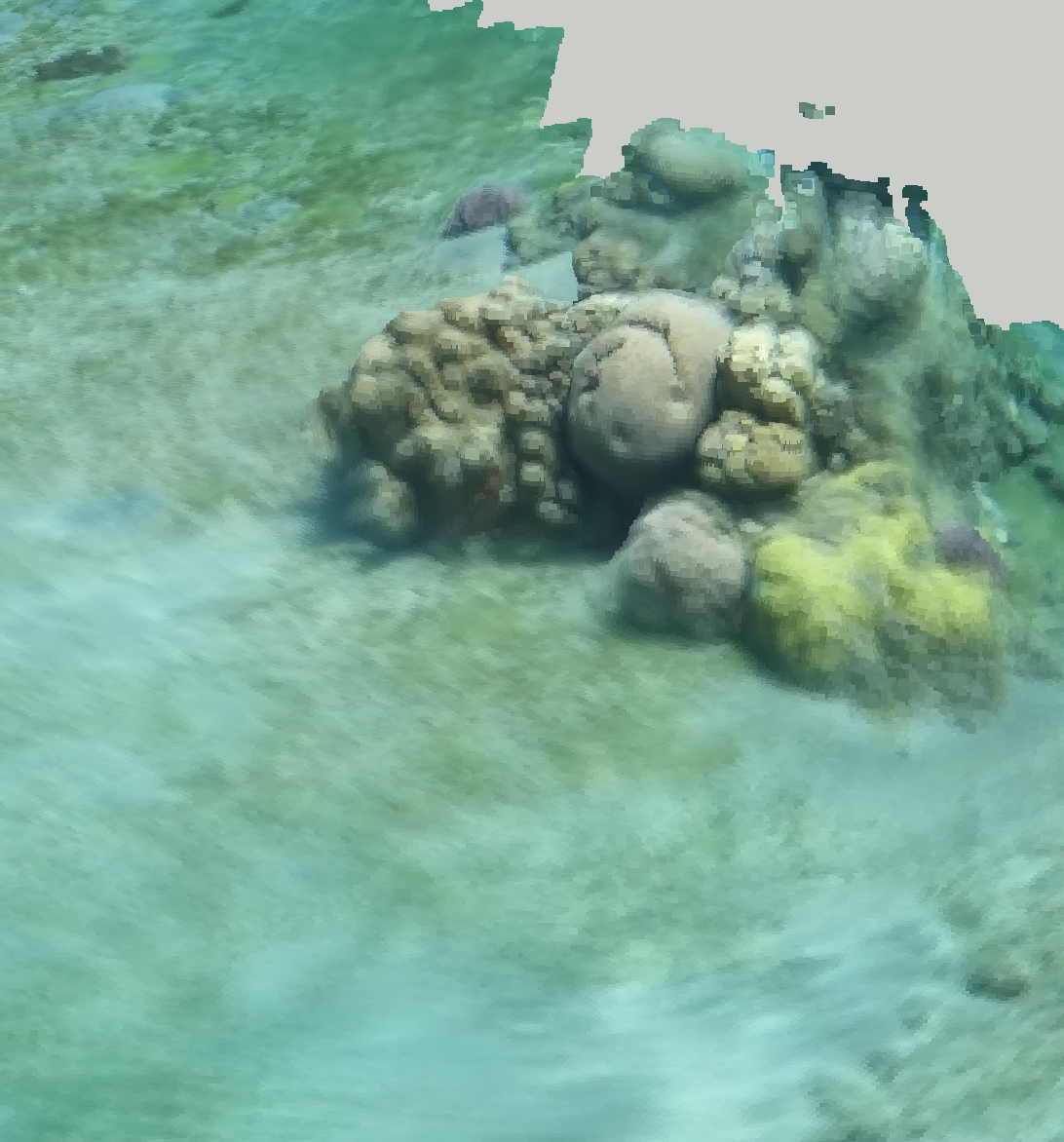}
		\includegraphics[height=88px]{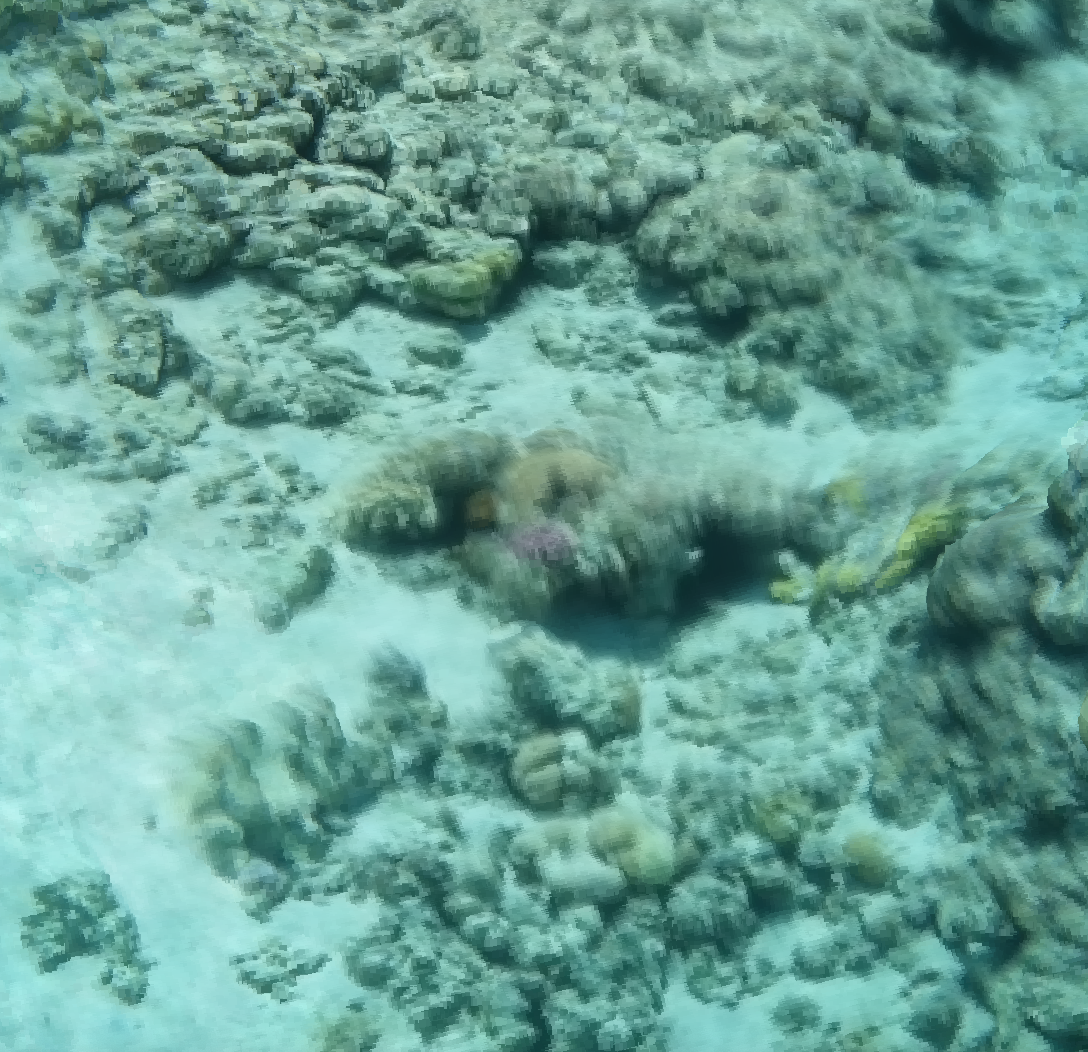}
		\vspace{-2pt}
		\caption{\new{Screenshots of various 3D reconstructions reef scenes from DeepReefMap. The top left screenshot is a close-up of the low-res top-down point cloud from Figure \ref{fig:spatial_comparison}, demonstrating local coherence even when global coherence is not recovered due to absence of gravity vectors from the camera's inertial measurement unit. The remaining images are from various ego-motion video scenes from the dataset.}}
		\label{fig:samples}
	\end{figure}

	\new{For a qualitative assessment, close-up screenshots from point clouds reconstructed by DeepReefMap are shown in Figure \ref{fig:samples}, demonstrating the level of detail that can be captured, and showing the diversity of scenes in which reconstruction  works reliably.}

	\begin{figure}[b!]
		\vspace{-3pt}
		\centering
		\begin{subfigure}[t]{\linewidth}
			\hfill
			\includegraphics[width=0.32\linewidth]{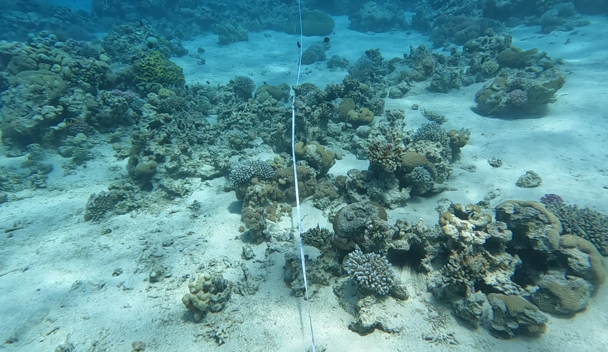} 
			\includegraphics[width=0.32\linewidth]{figures/rgbs_ka_rocky.png}
			\includegraphics[width=0.32\linewidth]{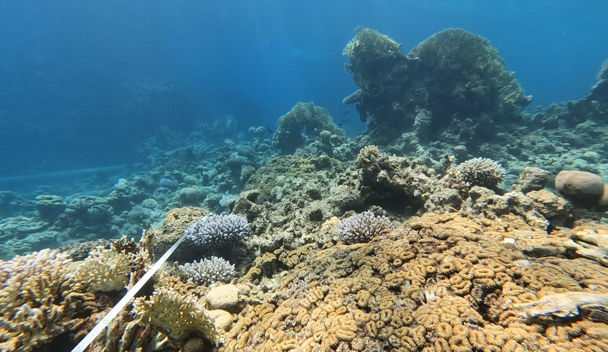}
			\hfill
			\caption{RGB Video Frames}
		\end{subfigure}
		
		\centering
		\begin{subfigure}[t]{\linewidth}
			\hfill
			\includegraphics[width=0.32\linewidth]{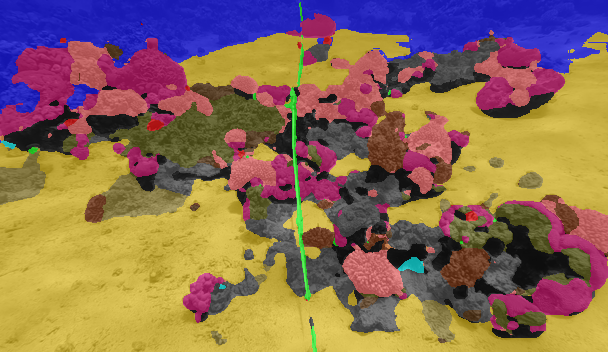}         
			\includegraphics[width=0.32\linewidth]{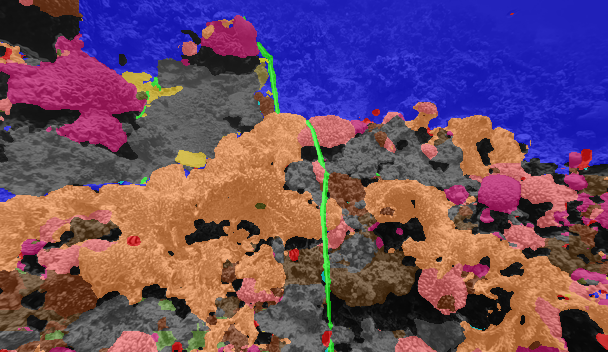}
			\includegraphics[width=0.32\linewidth]{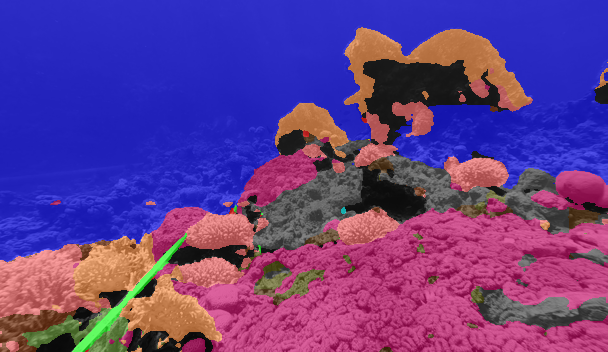}
			\hfill
			\caption{Neural Network Predictions}
		\end{subfigure}
		\centering
		
		\begin{subfigure}[t]{\linewidth}
			\hfill
			\includegraphics[width=0.32\linewidth]{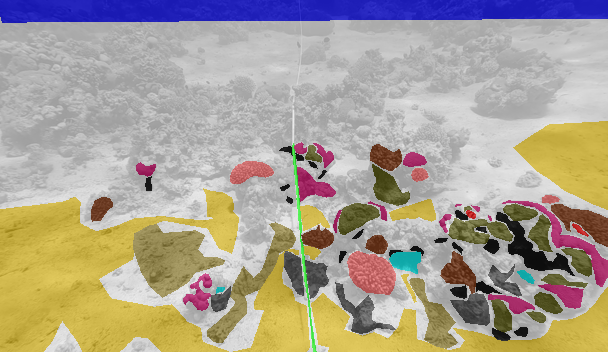}   
			\includegraphics[width=0.32\linewidth]{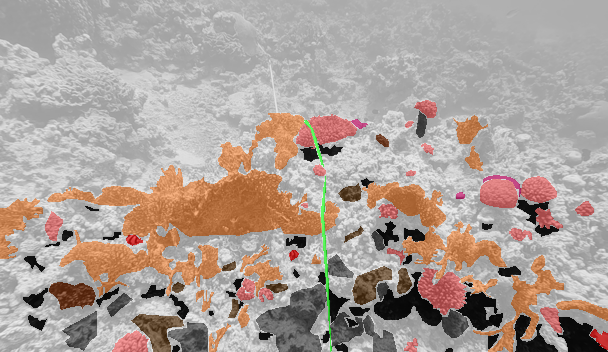}
			\includegraphics[width=0.32\linewidth]{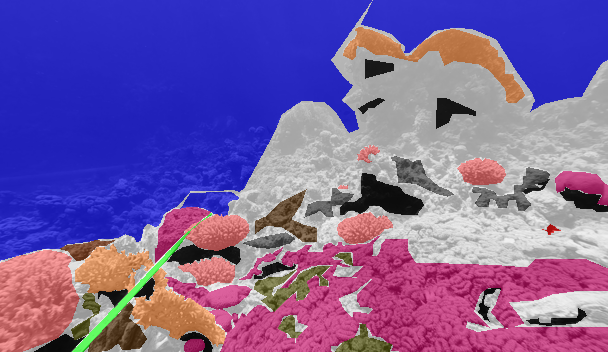}
			\hfill\\
			\vspace{1pt}
			\centering
			\includegraphics[width=0.95\linewidth]{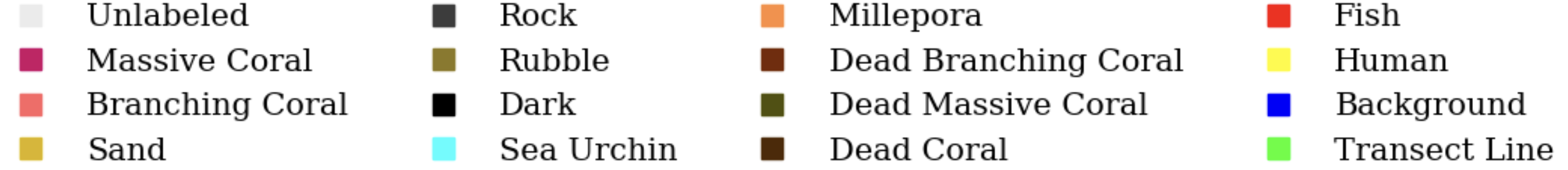}
			\caption{Ground Truth Annotations}
		\end{subfigure}
		
		\begin{subfigure}[t]{\linewidth}
			\hfill
			\includegraphics[width=0.32\linewidth]{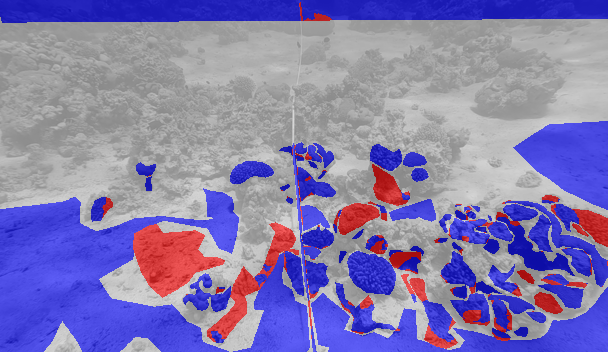}   
			\includegraphics[width=0.32\linewidth]{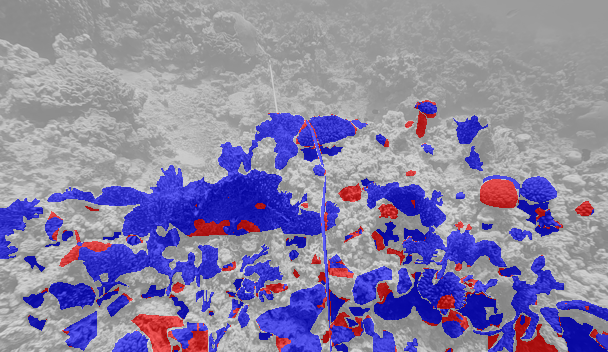}
			\includegraphics[width=0.32\linewidth]{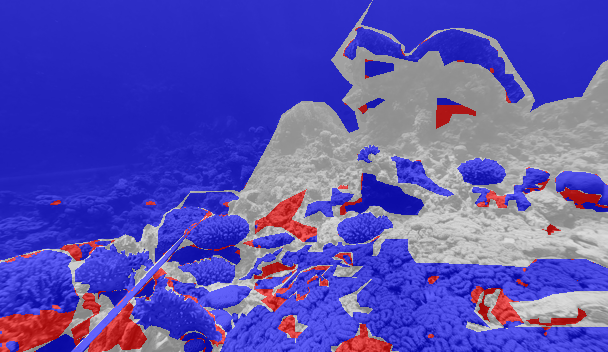}
			\hfill\\
			\centering
			\includegraphics[width=0.55\linewidth]{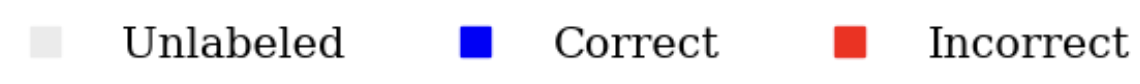}
			
			\caption{Correct Classifications}
		\end{subfigure}\vspace{-3pt}
		\caption{
			Example video frames from the three evaluation transects (top row), their respective benthic class predictions by the neural network (second row), the ground truth annotations (third row), and the mask of pixels which was labeled correctly and incorrectly (bottom row).
		}
		\label{fig:prediction}
	\end{figure}
	
	\new{\textbf{Semantic segmentation.} To evaluate the semantic segmentation, we start with a qualitative assessment:} we show video frames with their respective predicted benthic class, their ground-truth annotations, and the correctly and falsely predicted in Figure \ref{fig:prediction}. Most polygons are entirely classified correctly. The largest visible misclassifications come from assigning the wrong class to an entire large area, such misclassifying a patch of rubble as sand, or classifying a sea urchin tucked into a hole into the `dark' class.

	\begin{table*}[t!]
		\centering
		\caption{The accuracy over all pixels, the mean accuracy over all classes, the accuracy for each individual class appearing in the three evaluation transects, \new{and the number of pixels annotated in total and for each class in parentheses}. \new{Encrusting coral and soft coral did not visibly appear on the three evaluation transects.}}
		\resizebox{\textwidth}{!}{%
			\begin{tabular}{lccccccccccccccccccc}
				\toprule
				Transect & \makecell{\textbf{Total}\\ \textbf{Accuracy}} & \makecell{\textbf{Mean}\\ \textbf{Class}\\ \textbf{Accuracy}} & \makecell{Massive\\ Coral} & \makecell{Branching\\ Coral} & Sand & Rock & Rubble & Dark & \makecell{Sea\\ Urchin} & \makecell{Dead\\ Branching\\ Coral} & \makecell{Dead\\ Massive\\ Coral} & \makecell{Dead\\ Coral} & \makecell{Macro-\\algae} & Fish & Human & \makecell{Back-\\ground} & Trash & \makecell{Transect\\ Line}\\
				\toprule
				\midrule
				\makecell{King\\ Abdullah\\ (Sandy)} & \makecell{\textbf{86.65}\\(20653k)} & \textbf{65.83} & \makecell{94.10\\ (1272k)} & \makecell{91.73\\(391k)} & \makecell{98.36\\ (9303k)} & \makecell{66.19\\ (890k)} & \makecell{43.85\\ (2017k)} & \makecell{90.81\\ (793)} & \makecell{34.49\\ (29k)} &  \makecell{57.31\\ (744k)} & \makecell{62.18\\ (919k)} & \makecell{30.42\\ (222k)} & \makecell{-\\ (0)} & \makecell{72.71\\ (67k)} & \makecell{-\\ (0)} & \makecell{97.47\\ (3670k)} & \makecell{9.12\\ (27k)} & \makecell{81.97\\ (180k)}\\
				\midrule
				\makecell{King\\ Abdullah\\ (Rocky)} & \makecell{\textbf{79.84}\\(14078k)} & \textbf{70.08} & \makecell{91.21\\ (635k)} & \makecell{89.06\\ (432k)} & \makecell{91.40\\ (757k)} & \makecell{83.12\\ (2904k)} & \makecell{56.32\\ (2620k)} & \makecell{90.61\\ (1044k)} & \makecell{62.69\\ (129k)} &  \makecell{67.43\\ (426k)} & \makecell{26.01\\ (555k)} & \makecell{10.07\\ (280k)} & \makecell{-\\ (0)} & \makecell{79.98\\ (51k)} & \makecell{-\\ (0)} & \makecell{99.99\\ (3576k)} & \makecell{39.41\\ (37k)} & \makecell{93.84\\ (152k)}\\
				\midrule
				\makecell{Japanese\\ Garden\\ Israel}&\makecell{\textbf{84.35}\\(31114k)} & \textbf{69.13} & \makecell{93.65\\ (4843k)} & \makecell{81.61\\ (1490k)} & \makecell{98.04\\ (3674k)} & \makecell{81.03\\ (2890k)} & \makecell{53.19\\ (40k)} & \makecell{78.41\\ (1500k)} & \makecell{-\\ (0)} & \makecell{48.85\\ (835k)} & \makecell{34.67\\ (1464k)} & \makecell{6.604\\ (906k)} & \makecell{57.58\\ (433k)} & \makecell{77.82\\ (21k)} & \makecell{80.59\\ (207k)} & \makecell{94.36\\ (11948k)} & \makecell{-\\ (0)} & \makecell{81.43\\ (427k)}\\
				\bottomrule
		\end{tabular}}
		\label{table:results}
	\end{table*}
	
	\begin{figure*}[b!]
		\centering
		\begin{subfigure}[b]{0.497\linewidth}
			\centering
			\includegraphics[width=\linewidth]{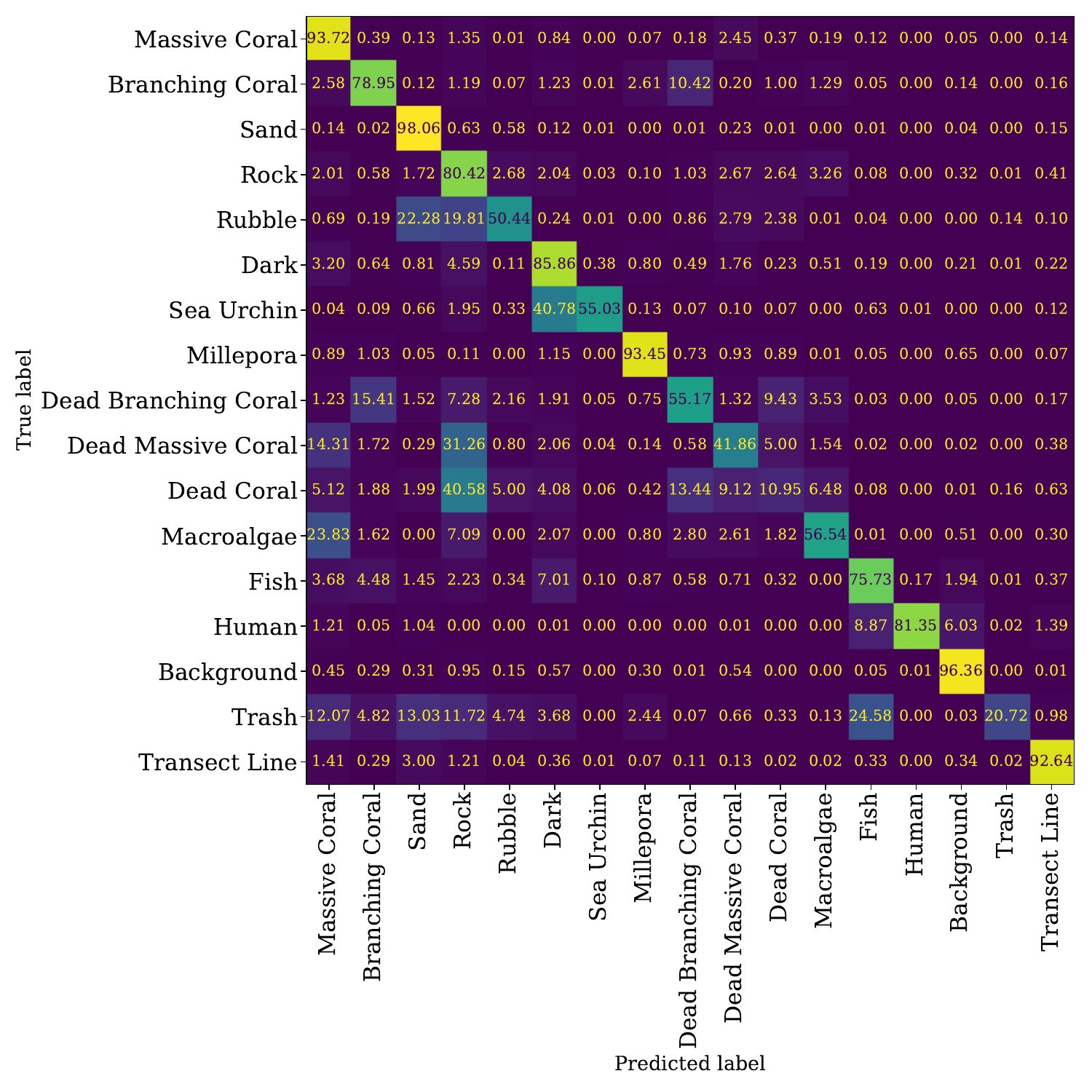}
			\caption{}
			\label{fig:confusiona}
		\end{subfigure}
		\hfill
		\begin{subfigure}[b]{0.497\linewidth}
			\centering
			\includegraphics[width=\linewidth]{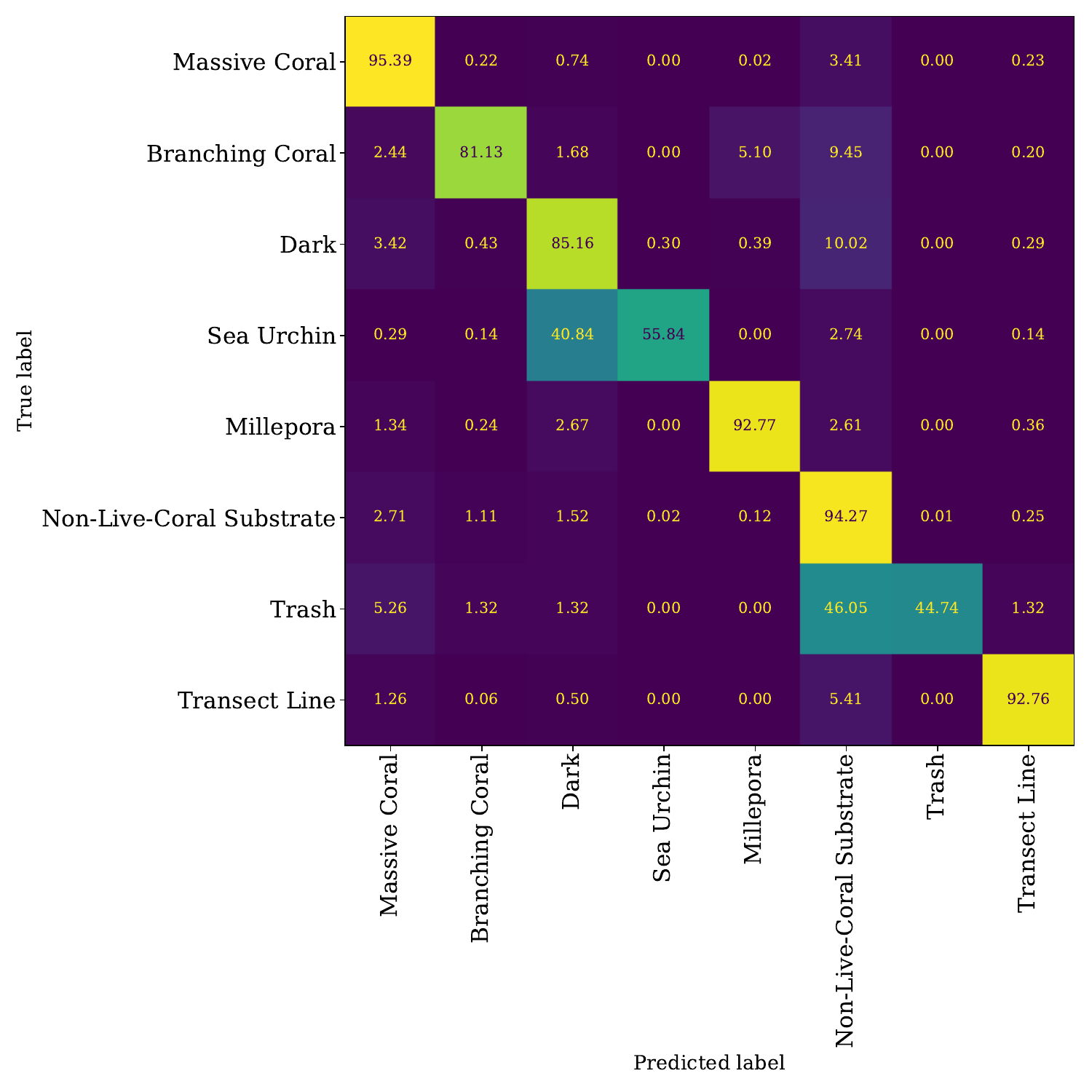}
			\caption{}
			\label{fig:confusionb}
		\end{subfigure}
		
		\caption{Confusion matrices on the three evaluation transects (a) evaluated at the image level with all classes present and (b) evaluated after projection on the point cloud level with all non-live-coral substrate classes summarized.}
		\label{fig:confusion}
	\end{figure*}
	
	\new{To quantify the accuracy of the semantic segmentation system}, we report the percentage of annotated pixels for which the neural network predicts the correct class as a metric. Computed accuracies for the individual label classes, as well as averaged over all classes and over all pixels are displayed in Table \ref{table:results}. All in all, the results are largely consistent for the three transects: the mean class accuracy is comparable between the three transects. Similarly, there is a large consistency between classes that are accurately classified: live massive coral, live branching coral, sand, transect line and sand, are all classified with at least 80\% accuracy on all three evaluation transects. On the other hand, classes with the lowest accuracy are consistently trash and dead coral. The total accuracy is higher than the mean class accuracy because classes that are easily segmented, such as background or sand, have more and larger annotated polygons. The main result is, that even with only a few hundred annotated image patches, the main benthic classes of interest can be detected with satisfactory accuracy using frame-wise dense segmentation neural networks.

	\begin{figure*}[t!]
		\centering
		\begin{subfigure}[b]{0.69\linewidth}
			\centering
			\includegraphics[width=\linewidth, height=118px]{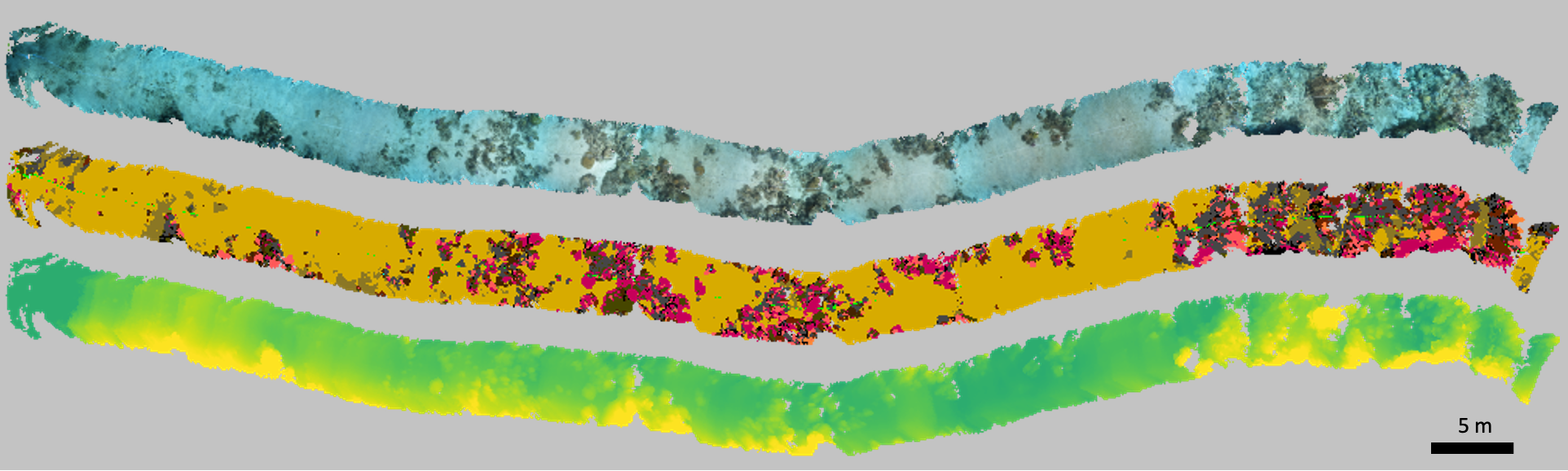}
			\caption{King Abdullah Reef (Sandy)}
		\end{subfigure}
		\hfill
		\begin{subfigure}[b]{0.3\linewidth}
			\raggedright
			\includegraphics[height=145px,width=\linewidth]{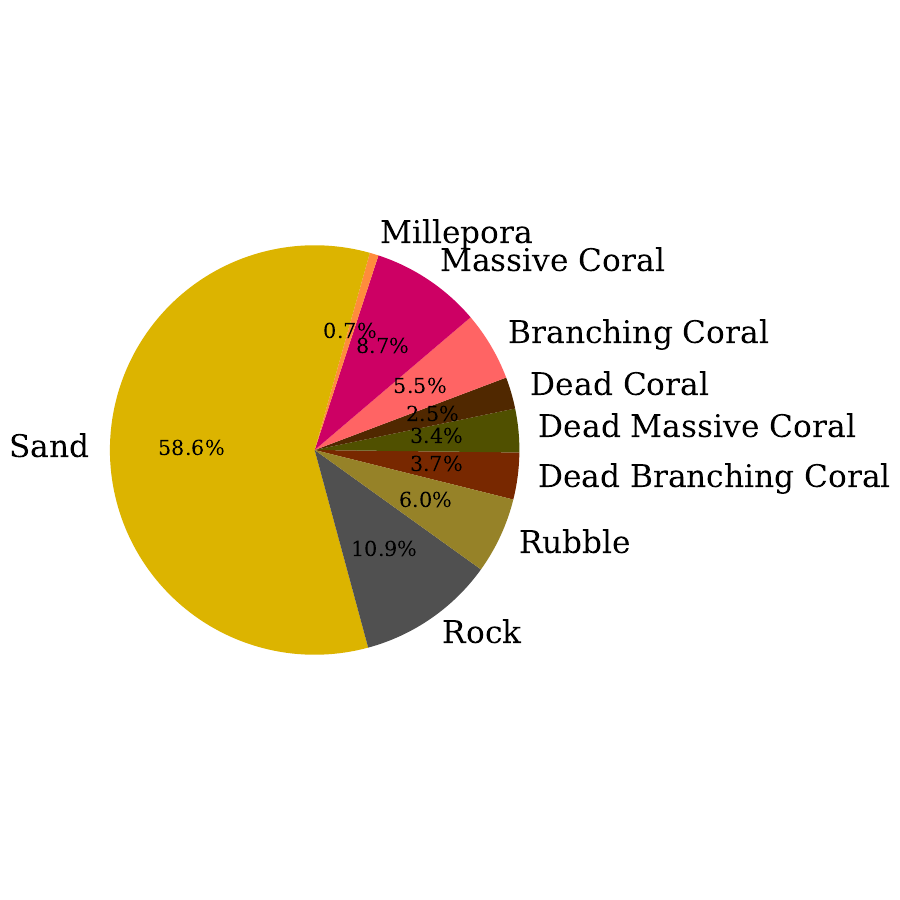}
		\end{subfigure}

		\centering
		\begin{subfigure}[b]{0.69\linewidth}
			\centering
			\includegraphics[width=\linewidth, height=118px]{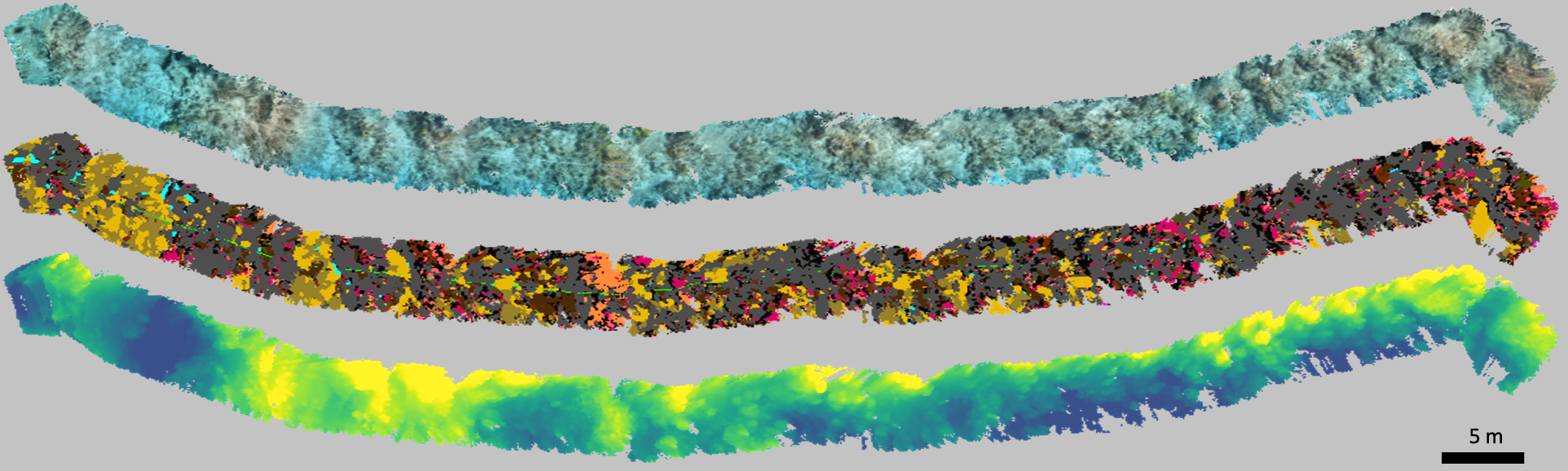}
			\caption{King Abdullah Reef (Rocky)}
		\end{subfigure}
		\hfill
		\begin{subfigure}[b]{0.3\linewidth}
			\raggedright
			\includegraphics[height=143px,width=0.96\linewidth]{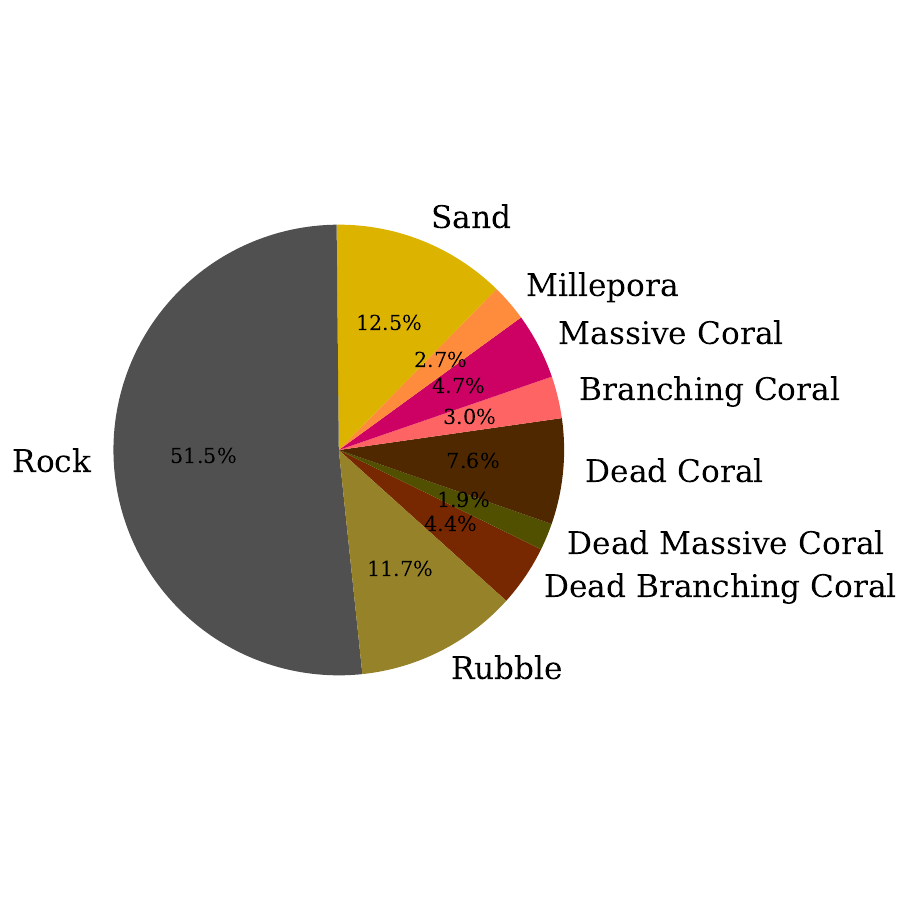}
		\end{subfigure}
		
		\centering
		\begin{subfigure}[b]{0.69\linewidth}
			\centering
			\includegraphics[width=\linewidth, height=118px]{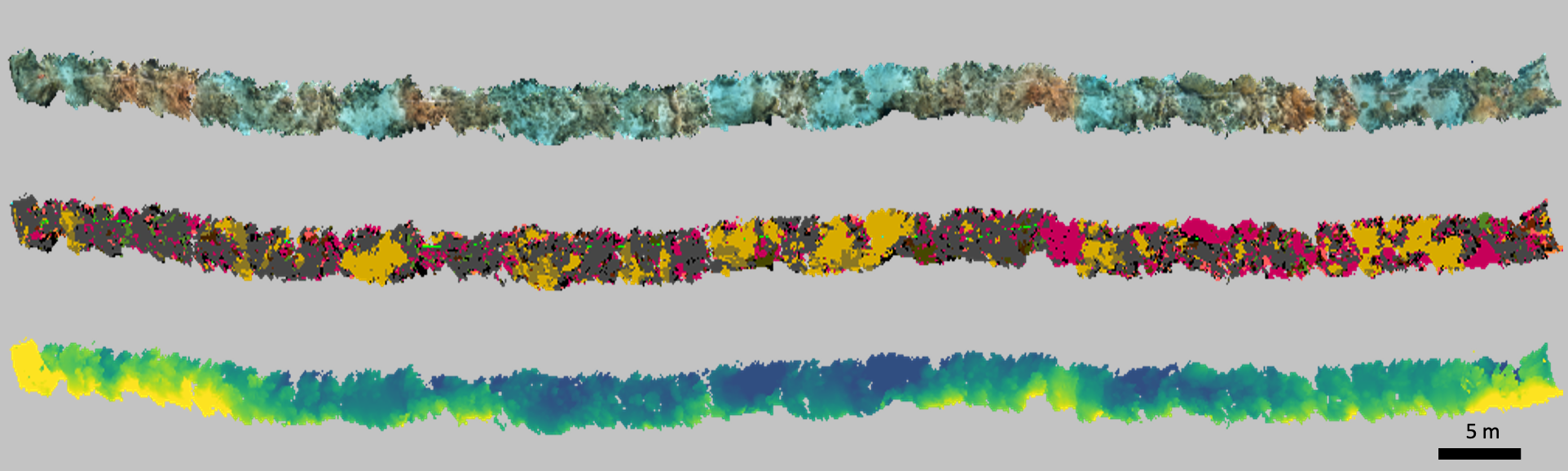}
			\includegraphics[width=0.6\linewidth]{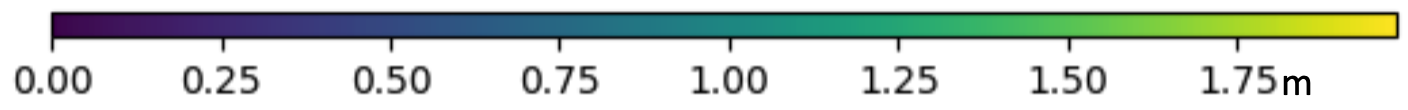}
			\caption{Japanese Garden Israel}
		\end{subfigure}
		\hfill
		\begin{subfigure}[b]{0.3\linewidth}
			\raggedright
			\includegraphics[height=125px,width=0.87\linewidth]{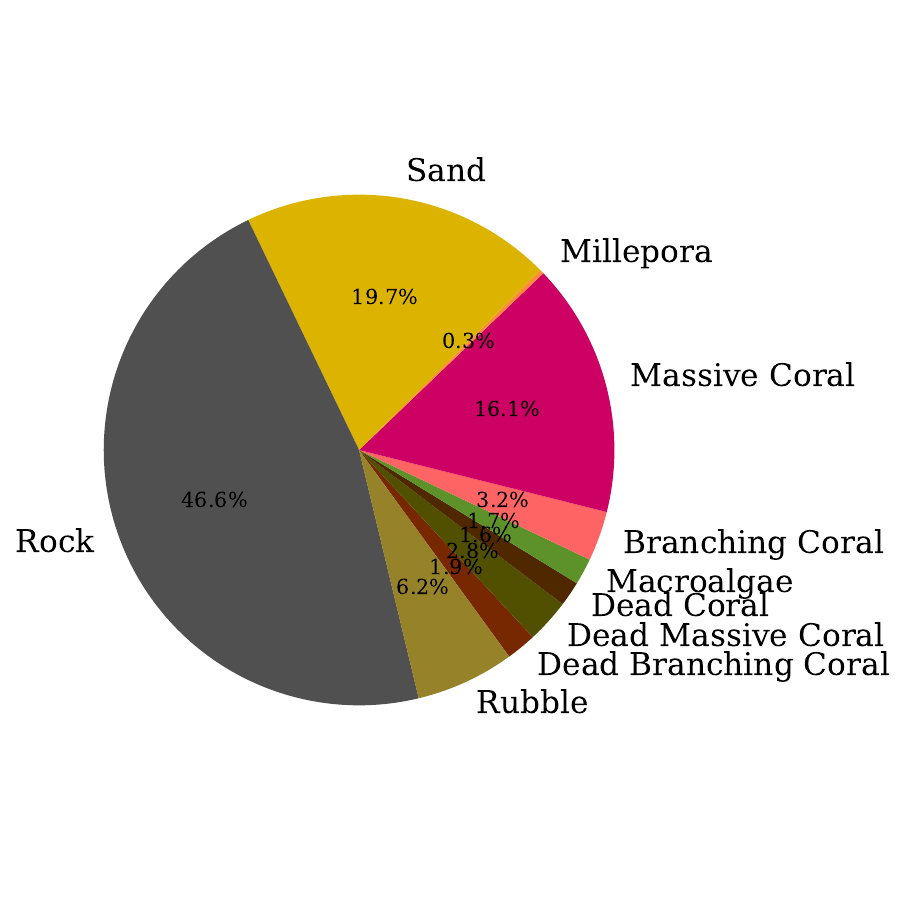}
			\vspace{25pt}
		\end{subfigure}

		\caption{Birds-eye view of the three evaluation transects of 100 m length with their RGB color, their semantic class, and their z-value (left), as well as a pie chart showing the percentage cover of the benthic substrate classes (right).}
		\label{fig:show_transects}
	\end{figure*}

	
	To better understand the misclassification errors on the image level, a confusion matrix of the union of the three evaluation transects is shown in Figure \ref{fig:confusiona}. Over the three transects in total, a mean class accuracy of 68.6\% is reached with 84.1\% of pixels being correctly classified. 
	
	For the `trash' class, which has the second lowest accuracy, many pixels are misclassified as fish, sand or rock. Marine litter is a strongly heterogeneous class: from aluminum cans that share bright colors with some fish, to pale-colored bits of cloth or plastic lying partially covered by sandy or rocky substrate. Most errors in the sea urchin and fish class are from misclassifications to `dark'. The majority of the remaining errors stem from the dead coral classes being confused with rock and rubble. To summarize, we find that a large portion of misclassifications are consistent with the challenges from the inherent ambiguity we faced while annotating the frames, as described above.

	\new{The ability} to distinguish the main classes of live coral from dead substrates in order to calculate the live coral cover is particularly important from an ecological perspective. This should be evaluated at the point cloud level, where unwanted classes and noisy or uncertain points have been removed, rather than at the image level. When rock, sand, rubble, macroalgae, and the dead coral classes are summarized into a non-live-coral substrate class, the mean class accuracy at the point cloud level is 80.2\%, with 92.6\% of points correctly classified. The remaining misclassifications are mostly from sea urchins being classified as dark, and trash being misclassified, as highlighted in Figure \ref{fig:confusionb}.

	We show ortho-projections of the 3D models of the three evaluation transects in their RGB color, colorized by their benthic class and their z-value, along with the area covered by each static benthic class in Figure \ref{fig:show_transects}, with a chosen grid cell size of 5 cm. The benthic cover percentages are computed straight from the ortho-projection by normalizing by the number of occupied 2D grid cells. In photogrammetry, occlusion within the image frames creates `shadows' in the point cloud, which can form holes in the 2D ortho-projection. For all three evaluated point clouds, holes that are completely enclosed in the 2D map account for less than 0.2\% of the area. This means that they have almost no impact on the benthic cover analysis. The width of the point clouds depends on the height of the diver above the seafloor and the angle of the camera for the individual videos. 
	
	\section*{Discussion}
	
	This paper presents a novel approach for \new{rapid} 3D semantic mapping of coral reefs from only ego-motion videos taken with a single action camera. This \new{approach significantly scales up the analysis of video transects. This} has broad implications on coral reef monitoring and may inspire new research directions for mapping underwater environments in general. This Section discusses limitations as well as main future developments that will ultimately determine the impact of the method.
	
	For coral reef ecology purposes, any analysis largely depends on the quality of the segmentation system. While our system can demonstrably separate the main classes of interest with a high accuracy, there is a large potential to improve the semantic segmentation to unlock a more fine-grained and accurate ecological analysis. In particular, for a more precise ecological analysis that is applicable to reef areas beyond the northern Red Sea, a much larger and more comprehensive dataset of semantic segmentation annotations must be created. This dataset should include scenes from more diverse reef environments, and annotations at the most precise taxonomic rank that is determinable from the images. With annotations down to the coral genus level and sufficiently large dataset size, opportunities for a higher level of sophistication in the deep learning process will become possible with more involved neural network architectures, training procedures, and the integration of information from multiple subsequent frames, which has the potential to lead to more precise semantic segmentation. 
	
	The main methodological contribution in this work introduces the paradigm of learning-based SfM to underwater ecosystems mapping, which promises to tackle the challenging conditions of underwater environments by learning from large video collections.
	From a computer vision perspective, the underlying methods are rapidly evolving: learning-based SfM methods are improving dramatically with novel neural network architectures and better loss formulations \citep{designdecisionsthatmatter} leading to better depth and pose estimates. Currently, while being significantly faster and more robust, learning-based SfM systems commonly do not reach the same accuracy as conventional SfM systems: global pose graph optimization is generally omitted in learning-based SfM and images fed into deep learning systems have to be significantly reduced in resolution due to limited GPU memory. Commonly, to match the final accuracy of conventional SfM, the estimated depths are scaled up using super-resolution procedures \citep{depthsuperresolution, droid} and subsequently fed into conventional SfM systems along with the pose transformation estimates to be fine-tuned, at a substantial computational cost  \citep{scaleconsistent, sfmlearner}. Nonetheless, while conventional SfM photogrammetry is an established research field dating back decades \citep{colmap}, the advances in learning-based SfM have been staggering, promising to tackle current limitations in the near future. To the best of our knowledge, no previously existing conventional SfM system can reliably deliver 3D reconstructions from \new{ego-motion} videos of reef scenes, especially not at a comparable computational cost.

	Furthermore, like all SfM systems, learning-based SfM suffers from the fact that tiny errors in the camera pose transformations behave multiplicatively: while the 3D reconstructions look coherent on a local scale, the global trajectory of the point cloud is commonly increasingly erroneous as the covered distance increases. For terrestrial systems, the remedy is given by precise GPS devices. Even so, detecting when the camera revisits a place in 3D space, detecting so called \textit{loop closures}, remains extremely challenging \citep{colmap, overlapnet}. Without an accurate positioning system under water, the described method is limited to 3D reconstructions without loop closures. Such positioning systems exist, but are orders of magnitude more expensive than the cameras we used, defeating the purpose of democratizing coral reef monitoring. In our proposed low-cost approach, just as in conventional coral reef monitoring methods, accurate geolocalization is thus only possible when markers with known GPS coordinates, such as transect lines, are visible. Nonetheless, as learning-based SfM methods evolve and improve, the effect of tiny inaccuracies in the pose transformations will become less detrimental.
	
	Another future work is the fusion of point clouds from multiple video mappings from the same areas, allowing to fill the 'shadows' in occluded views, and laying the foundation for more precise ecological analysis. This involve either merging the point clouds taken from cameras with different viewing angles, for example one front-facing and one rear-facing camera, or from multiple passes back and forth with one camera. Point clouds without holes from occlusion in them could, in principle, enable volumetric ecological analysis: beyond the areas covered by benthic classes, estimates for the biomass or the volume distribution of corals could be calculated.
	
	Without changing the underlying computational methods, our approach could be extended to other underwater environments, for example the submerged ecosystems of mangrove forests \citep{mangrove}, or to guide deep sea exploration \& salvage operations \citep{estonia}. With the availability of sizeable ego-motion video data from such scenarios, including a reasonably sized dataset of annotated video frames with the classes of interest, our approach should transfer seamlessly to such domains.
	
	\section*{Acknowledgments}
	We thank Dr. Ali Al-Sawalmih and Tariq Al-Salman (Marine Science Station, Aqaba), Prof. Maoz Fine, Nahum Sela (InterUniversity Institute of Marine Science, Eilat), Dr. Assaf Zvuloni (Nature Reserve Authority Israel), and the Aqaba Special Economic Zone Authority for their support for enabling us to collect the videos. Freya Behrens is thanked for her help on computational 3D geometry. The data of this study were collected in the framework of the Transnational Red Sea Center hosted by the Laboratory for Biological Geochemistry at EPFL. This work was funded in part by FNS grant 205321\_212614, as well as EPFL and the Transnational Red Sea Center.

	\section*{Author Contributions}
	Jonathan Sauder and Devis Tuia conceived the ideas and designed methodology; Jonathan Sauder, Guilhem Banc-Prandi and Anders Meibom collected the data; Jonathan Sauder wrote the source code and analyzed the data, and led the writing of the manuscript; all authors contributed substantially to writing and revising the manuscript and gave final approval for publication.
	
	\section*{Data Availability}
	
	Data is made available upon request. Upon final peer-reviewed publication, the data and source code will be made fully available.
	
	\bibliography{scifile}

\begin{thebibliography}{51}
\providecommand{\natexlab}[1]{#1}
\providecommand{\url}[1]{\texttt{#1}}
\expandafter\ifx\csname urlstyle\endcsname\relax
  \providecommand{\doi}[1]{doi: #1}\else
  \providecommand{\doi}{doi: \begingroup \urlstyle{rm}\Url}\fi

\bibitem[AgiSoft(2022)]{agisoft}
AgiSoft.
\newblock Metashape professional (version 1.8.3).
\newblock http://www.agisoft.com/downloads/installer/, 2022.

\bibitem[Alonso et~al.(2019)Alonso, Yuval, Eyal, Treibitz, and
  Murillo]{alonso2019coralseg}
Inigo Alonso, Matan Yuval, Gal Eyal, Tali Treibitz, and Ana~C Murillo.
\newblock Coralseg: Learning coral segmentation from sparse annotations.
\newblock \emph{Journal of Field Robotics}, 36\penalty0 (8):\penalty0
  1456--1477, 2019.

\bibitem[Asner et~al.(2020)Asner, Vaughn, Balzotti, Brodrick, and
  Heckler]{airborne}
Gregory~P Asner, Nicholas~R Vaughn, Christopher Balzotti, Philip~G Brodrick,
  and Joseph Heckler.
\newblock High-resolution reef bathymetry and coral habitat complexity from
  airborne imaging spectroscopy.
\newblock \emph{Remote Sensing}, 12\penalty0 (2):\penalty0 310, 2020.

\bibitem[Beijbom et~al.(2012)Beijbom, Edmunds, Kline, Mitchell, and
  Kriegman]{coralnet}
Oscar Beijbom, Peter~J Edmunds, David~I Kline, B~Greg Mitchell, and David
  Kriegman.
\newblock Automated annotation of coral reef survey images.
\newblock In \emph{2012 IEEE conference on computer vision and pattern
  recognition}, pages 1170--1177. IEEE, 2012.

\bibitem[Beijbom et~al.(2015)Beijbom, Edmunds, Roelfsema, Smith, Kline, Neal,
  Dunlap, Moriarty, Fan, Tan, et~al.]{svmcoral}
Oscar Beijbom, Peter~J Edmunds, Chris Roelfsema, Jennifer Smith, David~I Kline,
  Benjamin~P Neal, Matthew~J Dunlap, Vincent Moriarty, Tung-Yung Fan, Chih-Jui
  Tan, et~al.
\newblock Towards automated annotation of benthic survey images: Variability of
  human experts and operational modes of automation.
\newblock \emph{PloS one}, 10\penalty0 (7):\penalty0 e0130312, 2015.

\bibitem[Beyer et~al.(2018)Beyer, Kennedy, Beger, Chen, Cinner, Darling, Eakin,
  Gates, Heron, Knowlton, et~al.]{beyer2018risk}
Hawthorne~L Beyer, Emma~V Kennedy, Maria Beger, Chaolun~Allen Chen, Joshua~E
  Cinner, Emily~S Darling, C~Mark Eakin, Ruth~D Gates, Scott~F Heron, Nancy
  Knowlton, et~al.
\newblock Risk-sensitive planning for conserving coral reefs under rapid
  climate change.
\newblock \emph{Conservation Letters}, 11\penalty0 (6):\penalty0 e12587, 2018.

\bibitem[Bian et~al.(2021)Bian, Zhan, Wang, Li, Zhang, Shen, Cheng, and
  Reid]{scaleconsistent}
Jia-Wang Bian, Huangying Zhan, Naiyan Wang, Zhichao Li, Le~Zhang, Chunhua Shen,
  Ming-Ming Cheng, and Ian Reid.
\newblock Unsupervised scale-consistent depth learning from video.
\newblock \emph{International Journal of Computer Vision}, 129\penalty0
  (9):\penalty0 2548--2564, 2021.

\bibitem[Bongaerts et~al.(2021)Bongaerts, Dub{\'e}, Prata, Gijsbers, Achlatis,
  and Hernandez-Agreda]{reefscapegenomics}
Pim Bongaerts, Caroline~E Dub{\'e}, Katharine~E Prata, Johanna~C Gijsbers,
  Michelle Achlatis, and Alejandra Hernandez-Agreda.
\newblock Reefscape genomics: leveraging advances in 3d imaging to assess
  fine-scale patterns of genomic variation on coral reefs.
\newblock \emph{Frontiers in Marine Science}, page 875, 2021.

\bibitem[Burns et~al.(2015)Burns, Delparte, Gates, and Takabayashi]{burns}
JHR Burns, D~Delparte, RD~Gates, and M~Takabayashi.
\newblock Integrating structure-from-motion photogrammetry with geospatial
  software as a novel technique for quantifying 3d ecological characteristics
  of coral reefs.
\newblock \emph{PeerJ}, 3:\penalty0 e1077, 2015.

\bibitem[Carleton and Done(1995)]{videotransect}
JH~Carleton and TJ~Done.
\newblock Quantitative video sampling of coral reef benthos: large-scale
  application.
\newblock \emph{Coral Reefs}, 14:\penalty0 35--46, 1995.

\bibitem[Chen et~al.(2021{\natexlab{a}})Chen, Beijbom, Chan, Bouwmeester, and
  Kriegman]{coralnetengine}
Qimin Chen, Oscar Beijbom, Stephen Chan, Jessica Bouwmeester, and David
  Kriegman.
\newblock A new deep learning engine for coralnet.
\newblock In \emph{Proceedings of the IEEE/CVF International Conference on
  Computer Vision}, pages 3693--3702, 2021{\natexlab{a}}.

\bibitem[Chen et~al.(2021{\natexlab{b}})Chen, L{\"a}be, Milioto, R{\"o}hling,
  Vysotska, Haag, Behley, and Stachniss]{overlapnet}
Xieyuanli Chen, Thomas L{\"a}be, Andres Milioto, Timo R{\"o}hling, Olga
  Vysotska, Alexandre Haag, Jens Behley, and Cyrill Stachniss.
\newblock Overlapnet: Loop closing for lidar-based slam.
\newblock \emph{arXiv preprint arXiv:2105.11344}, 2021{\natexlab{b}}.

\bibitem[Chennu et~al.(2017)Chennu, F{\"a}rber, De’ath, De~Beer, and
  Fabricius]{hyperdiver}
Arjun Chennu, Paul F{\"a}rber, Glenn De’ath, Dirk De~Beer, and Katharina~E
  Fabricius.
\newblock A diver-operated hyperspectral imaging and topographic surveying
  system for automated mapping of benthic habitats.
\newblock \emph{Scientific reports}, 7\penalty0 (1):\penalty0 7122, 2017.

\bibitem[Curless and Levoy(1996)]{tsdf}
Brian Curless and Marc Levoy.
\newblock A volumetric method for building complex models from range images.
\newblock In \emph{Proceedings of the 23rd annual conference on Computer
  graphics and interactive techniques}, pages 303--312, 1996.

\bibitem[Deng et~al.(2009)Deng, Dong, Socher, Li, Li, and Fei-Fei]{imagenet}
Jia Deng, Wei Dong, Richard Socher, Li-Jia Li, Kai Li, and Li~Fei-Fei.
\newblock Imagenet: A large-scale hierarchical image database.
\newblock In \emph{2009 IEEE conference on computer vision and pattern
  recognition}, pages 248--255. Ieee, 2009.

\bibitem[Dixon et~al.(2022)Dixon, Forster, Heron, Stoner, and Beger]{future}
Adele~M Dixon, Piers~M Forster, Scott~F Heron, Anne~MK Stoner, and Maria Beger.
\newblock Future loss of local-scale thermal refugia in coral reef ecosystems.
\newblock \emph{Plos Climate}, 1\penalty0 (2):\penalty0 e0000004, 2022.

\bibitem[Fine et~al.(2013)Fine, Gildor, and Genin]{maoz}
Maoz Fine, Hezi Gildor, and Amatzia Genin.
\newblock A coral reef refuge in the red sea.
\newblock \emph{Global change biology}, 19\penalty0 (12):\penalty0 3640--3647,
  2013.

\bibitem[Fisher et~al.(2015)Fisher, O’Leary, Low-Choy, Mengersen, Knowlton,
  Brainard, and Caley]{blabla}
Rebecca Fisher, Rebecca~A O’Leary, Samantha Low-Choy, Kerrie Mengersen, Nancy
  Knowlton, Russell~E Brainard, and M~Julian Caley.
\newblock Species richness on coral reefs and the pursuit of convergent global
  estimates.
\newblock \emph{Current Biology}, 25\penalty0 (4):\penalty0 500--505, 2015.

\bibitem[Giardino et~al.(2015)Giardino, Bresciani, Fava, Matta, Brando, and
  Colombo]{mangrove}
Claudia Giardino, Mariano Bresciani, Francesco Fava, Erica Matta, Vittorio~E
  Brando, and Roberto Colombo.
\newblock Mapping submerged habitats and mangroves of lampi island marine
  national park (myanmar) from in situ and satellite observations.
\newblock \emph{Remote Sensing}, 8\penalty0 (1):\penalty0 2, 2015.

\bibitem[He et~al.(2016)He, Zhang, Ren, and Sun]{resnet}
Kaiming He, Xiangyu Zhang, Shaoqing Ren, and Jian Sun.
\newblock Deep residual learning for image recognition.
\newblock In \emph{Proceedings of the IEEE conference on computer vision and
  pattern recognition}, pages 770--778, 2016.

\bibitem[Hopkinson et~al.(2020)Hopkinson, King, Owen, Johnson-Roberson, Long,
  and Bhandarkar]{reverse}
Brian~M Hopkinson, Andrew~C King, Daniel~P Owen, Matthew Johnson-Roberson,
  Matthew~H Long, and Suchendra~M Bhandarkar.
\newblock Automated classification of three-dimensional reconstructions of
  coral reefs using convolutional neural networks.
\newblock \emph{PloS one}, 15\penalty0 (3):\penalty0 e0230671, 2020.

\bibitem[Hughes et~al.(2017)Hughes, Kerry, {\'A}lvarez-Noriega,
  {\'A}lvarez-Romero, Anderson, Baird, Babcock, Beger, Bellwood, Berkelmans,
  et~al.]{hughes2017global}
Terry~P Hughes, James~T Kerry, Mariana {\'A}lvarez-Noriega, Jorge~G
  {\'A}lvarez-Romero, Kristen~D Anderson, Andrew~H Baird, Russell~C Babcock,
  Maria Beger, David~R Bellwood, Ray Berkelmans, et~al.
\newblock Global warming and recurrent mass bleaching of corals.
\newblock \emph{Nature}, 543\penalty0 (7645):\penalty0 373--377, 2017.

\bibitem[Jakobsson et~al.(2021)Jakobsson, Stranne, Fornander, O'Regan, and
  Wagner]{estonia}
Martin Jakobsson, Christian Stranne, Rickard Fornander, Matt O'Regan, and Anton
  Wagner.
\newblock El21-estonia: Report of the ms estonia shipwreck site survey with rv
  electra, 2021.

\bibitem[Kellenberger et~al.(2020)Kellenberger, Tuia, and Morris]{aide}
Benjamin Kellenberger, Devis Tuia, and Dan Morris.
\newblock Aide: Accelerating image-based ecological surveys with interactive
  machine learning.
\newblock \emph{Methods in Ecology and Evolution}, 11\penalty0 (12):\penalty0
  1716--1727, 2020.

\bibitem[Kingma and Ba(2015)]{adam}
Diederik~P Kingma and Jimmy Ba.
\newblock Adam: A method for stochastic optimization.
\newblock In \emph{International Conference on Learning Representations
  (ICLR)}, 2015.

\bibitem[Knowlton and Jackson(2008)]{knowlton2008shifting}
Nancy Knowlton and Jeremy B~C Jackson.
\newblock Shifting baselines, local impacts, and global change on coral reefs.
\newblock \emph{PLoS biology}, 6\penalty0 (2):\penalty0 e54, 2008.

\bibitem[Krueger et~al.(2017)Krueger, Horwitz, Bodin, Giovani, Escrig, Meibom,
  and Fine]{kruegerredsea}
Thomas Krueger, Noa Horwitz, Julia Bodin, Maria-Evangelia Giovani, St{\'e}phane
  Escrig, Anders Meibom, and Maoz Fine.
\newblock Common reef-building coral in the northern red sea resistant to
  elevated temperature and acidification.
\newblock \emph{Royal Society open science}, 4\penalty0 (5):\penalty0 170038,
  2017.

\bibitem[Leon et~al.(2015)Leon, Roelfsema, Saunders, and Phinn]{leon}
Javier~X Leon, Chris~M Roelfsema, Megan~I Saunders, and Stuart~R Phinn.
\newblock Measuring coral reef terrain roughness using
  ‘structure-from-motion’close-range photogrammetry.
\newblock \emph{Geomorphology}, 242:\penalty0 21--28, 2015.

\bibitem[Martin et~al.(2022)Martin, Russell, Hadfield, and
  Bowden]{designdecisionsthatmatter}
Jaime~Spencer Martin, Chris Russell, Simon Hadfield, and Richard Bowden.
\newblock Deconstructing self-supervised monocular reconstruction: The design
  decisions that matter.
\newblock \emph{arXiv preprint arXiv:2208.01489}, 2022.

\bibitem[Masson-Delmotte et~al.(2018)Masson-Delmotte, Zhai, P{\"o}rtner,
  Roberts, Skea, Shukla, Pirani, Moufouma-Okia, P{\'e}an, Pidcock,
  et~al.]{ipcc15}
Val{\'e}rie Masson-Delmotte, Panmao Zhai, Hans-Otto P{\"o}rtner, Debra Roberts,
  Jim Skea, Priyadarshi~R Shukla, Anna Pirani, Wilfran Moufouma-Okia, Clotilde
  P{\'e}an, Roz Pidcock, et~al.
\newblock Global warming of 1.5 c.
\newblock \emph{An IPCC Special Report on the impacts of global warming of},
  1\penalty0 (5), 2018.

\bibitem[Masson-Delmotte et~al.(2021)Masson-Delmotte, Zhai, Pirani, Connors,
  P{\'e}an, Berger, Caud, Chen, Goldfarb, Gomis, et~al.]{ipcc21}
Val{\'e}rie Masson-Delmotte, Panmao Zhai, Anna Pirani, Sarah~L Connors,
  Clotilde P{\'e}an, S~Berger, N~Caud, Y~Chen, L~Goldfarb, MI~Gomis, et~al.
\newblock Climate change 2021: the physical science basis.
\newblock \emph{Contribution of working group I to the sixth assessment report
  of the intergovernmental panel on climate change}, page~2, 2021.

\bibitem[NOAA(2022)]{halfabillion}
NOAA.
\newblock Fact sheet: Coral reefs.
\newblock \url{https://www.coast.noaa.gov/states/fast-facts/coral-reefs.html},
  2022.
\newblock Accessed: 2022-09-09.

\bibitem[Obura et~al.(2019)Obura, Aeby, Amornthammarong, Appeltans, Bax,
  Bishop, Brainard, Chan, Fletcher, Gordon, et~al.]{conventional}
David~O Obura, Greta Aeby, Natchanon Amornthammarong, Ward Appeltans, Nicholas
  Bax, Joe Bishop, Russell~E Brainard, Samuel Chan, Pamela Fletcher, Timothy~AC
  Gordon, et~al.
\newblock Coral reef monitoring, reef assessment technologies, and
  ecosystem-based management.
\newblock \emph{Frontiers in Marine Science}, 6:\penalty0 580, 2019.

\bibitem[Osman et~al.(2018)Osman, Smith, Ziegler, K{\"u}rten, Conrad,
  El-Haddad, Voolstra, and Suggett]{osmanredsea}
Eslam~O Osman, David~J Smith, Maren Ziegler, Benjamin K{\"u}rten, Constanze
  Conrad, Khaled~M El-Haddad, Christian~R Voolstra, and David~J Suggett.
\newblock Thermal refugia against coral bleaching throughout the northern red
  sea.
\newblock \emph{Global change biology}, 24\penalty0 (2):\penalty0 e474--e484,
  2018.

\bibitem[Prasil~Delaval et~al.(2021)Prasil~Delaval, Wicquart, Staub, and
  Planes]{icriconventional}
Nina Prasil~Delaval, Jeremy Wicquart, Francis Staub, and Serge Planes.
\newblock Status of coral reef monitoring: An assessment of methods and data at
  the national level.
\newblock \emph{International Coral Reef Initiative}, 2021.

\bibitem[Raoult et~al.(2016)Raoult, David, Dupont, Mathewson, O’Neill,
  Powell, and Williamson]{raoult2016gopros}
Vincent Raoult, Peter~A David, Sally~F Dupont, Ciaran~P Mathewson, Samuel~J
  O’Neill, Nicholas~N Powell, and Jane~E Williamson.
\newblock Gopros™ as an underwater photogrammetry tool for citizen science.
\newblock \emph{PeerJ}, 4:\penalty0 e1960, 2016.

\bibitem[Raoult et~al.(2017)Raoult, Reid-Anderson, Ferri, and Williamson]{sfm1}
Vincent Raoult, Sarah Reid-Anderson, Andreas Ferri, and Jane~E Williamson.
\newblock How reliable is structure from motion (sfm) over time and between
  observers? a case study using coral reef bommies.
\newblock \emph{Remote Sensing}, 9\penalty0 (7):\penalty0 740, 2017.

\bibitem[Ronneberger et~al.(2015)Ronneberger, Fischer, and Brox]{unet}
Olaf Ronneberger, Philipp Fischer, and Thomas Brox.
\newblock U-net: Convolutional networks for biomedical image segmentation.
\newblock In \emph{International Conference on Medical image computing and
  computer-assisted intervention}, pages 234--241. Springer, 2015.

\bibitem[Savary et~al.(2021)Savary, Barshis, Voolstra, C{\'a}rdenas, Evensen,
  Banc-Prandi, Fine, and Meibom]{savaryredsea}
Romain Savary, Daniel~J Barshis, Christian~R Voolstra, Anny C{\'a}rdenas,
  Nicolas~R Evensen, Guilhem Banc-Prandi, Maoz Fine, and Anders Meibom.
\newblock Fast and pervasive transcriptomic resilience and acclimation of
  extremely heat-tolerant coral holobionts from the northern red sea.
\newblock \emph{Proceedings of the National Academy of Sciences}, 118\penalty0
  (19):\penalty0 e2023298118, 2021.

\bibitem[Sch\"{o}nberger and Frahm(2016)]{colmap}
Johannes~Lutz Sch\"{o}nberger and Jan-Michael Frahm.
\newblock Structure-from-motion revisited.
\newblock In \emph{Conference on Computer Vision and Pattern Recognition
  (CVPR)}, 2016.

\bibitem[Sch{\"u}rholz and Chennu(2023)]{digitizing}
Daniel Sch{\"u}rholz and Arjun Chennu.
\newblock Digitizing the coral reef: Machine learning of underwater spectral
  images enables dense taxonomic mapping of benthic habitats.
\newblock \emph{Methods in Ecology and Evolution}, 14\penalty0 (2):\penalty0
  596--613, 2023.

\bibitem[Souter et~al.(2021)Souter, Planes, Wicquart, Logan, Obura, and
  Staub]{icrireport}
David Souter, Serge Planes, Jeremy Wicquart, Murray Logan, David Obura, and
  Francis Staub.
\newblock Status of coral reefs of the world: 2020.
\newblock \emph{Global Coral Reef Monitoring Network}, 2021.

\bibitem[Storlazzi et~al.(2016)Storlazzi, Dartnell, Hatcher, and
  Gibbs]{storlazzi}
Curt~D Storlazzi, Peter Dartnell, Gerald~A Hatcher, and Ann~E Gibbs.
\newblock End of the chain? rugosity and fine-scale bathymetry from existing
  underwater digital imagery using structure-from-motion (sfm) technology.
\newblock \emph{Coral Reefs}, 35\penalty0 (3):\penalty0 889--894, 2016.

\bibitem[Teed and Deng(2020)]{depthsuperresolution}
Zachary Teed and Jia Deng.
\newblock Raft: Recurrent all-pairs field transforms for optical flow.
\newblock In \emph{European conference on computer vision}, pages 402--419.
  Springer, 2020.

\bibitem[Teed and Deng(2021)]{droid}
Zachary Teed and Jia Deng.
\newblock Droid-slam: Deep visual slam for monocular, stereo, and rgb-d
  cameras.
\newblock \emph{Advances in Neural Information Processing Systems},
  34:\penalty0 16558--16569, 2021.

\bibitem[Urbina-Barreto et~al.(2021)Urbina-Barreto, Garnier, Elise, Pinel,
  Dumas, Mahamadaly, Facon, Bureau, Peignon, Quod, et~al.]{orthomosaic}
Isabel Urbina-Barreto, R{\'e}mi Garnier, Simon Elise, Romain Pinel, Pascal
  Dumas, Vincent Mahamadaly, Mathilde Facon, Sophie Bureau, Christophe Peignon,
  Jean-Pascal Quod, et~al.
\newblock Which method for which purpose? a comparison of line intercept
  transect and underwater photogrammetry methods for coral reef surveys.
\newblock \emph{Frontiers in Marine Science}, 8:\penalty0 636902, 2021.

\bibitem[Voolstra et~al.(2021)Voolstra, Valenzuela, Turkarslan, C{\'a}rdenas,
  Hume, Perna, Buitrago-L{\'o}pez, Rowe, Orellana, Baliga, et~al.]{voolstra}
Christian~R Voolstra, Jacob~J Valenzuela, Serdar Turkarslan, Anny C{\'a}rdenas,
  Benjamin~CC Hume, Gabriela Perna, Carol Buitrago-L{\'o}pez, Katherine Rowe,
  Monica~V Orellana, Nitin~S Baliga, et~al.
\newblock Contrasting heat stress response patterns of coral holobionts across
  the red sea suggest distinct mechanisms of thermal tolerance.
\newblock \emph{Molecular ecology}, 30\penalty0 (18):\penalty0 4466--4480,
  2021.

\bibitem[Williams et~al.(2019)Williams, Couch, Beijbom, Oliver, Vargas-Angel,
  Schumacher, and Brainard]{coralnetvsreal}
Ivor~D Williams, Courtney~S Couch, Oscar Beijbom, Thomas~A Oliver, Bernardo
  Vargas-Angel, Brett~D Schumacher, and Russell~E Brainard.
\newblock Leveraging automated image analysis tools to transform our capacity
  to assess status and trends of coral reefs.
\newblock \emph{Frontiers in Marine Science}, page 222, 2019.

\bibitem[Xie et~al.(2017)Xie, Girshick, Doll{\'a}r, Tu, and He]{resnext}
Saining Xie, Ross Girshick, Piotr Doll{\'a}r, Zhuowen Tu, and Kaiming He.
\newblock Aggregated residual transformations for deep neural networks.
\newblock In \emph{Proceedings of the IEEE conference on computer vision and
  pattern recognition}, pages 1492--1500, 2017.

\bibitem[Yuval et~al.(2021)Yuval, Alonso, Eyal, Tchernov, Loya, Murillo, and
  Treibitz]{yuval2021repeatable}
Matan Yuval, I{\~n}igo Alonso, Gal Eyal, Dan Tchernov, Yossi Loya, Ana~C
  Murillo, and Tali Treibitz.
\newblock Repeatable semantic reef-mapping through photogrammetry and
  label-augmentation.
\newblock \emph{Remote Sensing}, 13\penalty0 (4):\penalty0 659, 2021.

\bibitem[Zhou et~al.(2017)Zhou, Brown, Snavely, and Lowe]{sfmlearner}
Tinghui Zhou, Matthew Brown, Noah Snavely, and David~G Lowe.
\newblock Unsupervised learning of depth and ego-motion from video.
\newblock In \emph{Proceedings of the IEEE conference on computer vision and
  pattern recognition}, pages 1851--1858, 2017.

\end{thebibliography}
	
	\bibliographystyle{plainnat}

	\hfill
	\clearpage
	
	\subsection*{Appendix: Data}
	\vspace{-5pt}
	\begin{table}[ht]
		\centering
		\caption{Overview of the ego-motion video dataset from the Red Sea used to train the learning-based SfM system and the semantic segmentation system}
		\begin{tabular}{lccccc}
			\toprule
			Country & Site & \makecell{Video \\Length} & \makecell{Video \\ Frames} & \makecell{ Annotated \\Frame \\Patches} & \makecell{Annotated \\Pixels}\\
			\toprule
			\midrule
			\makecell{Israel} & \makecell{Japanese Garden} & 8h53 & 640k & 821 & 165.9M \\
			\midrule
			\makecell{Jordan} & \makecell{MSS Reef} & 3h37 & 260k & 94 & 11.4M \\
			\midrule
			\makecell{Jordan} & \makecell{South Port Phosphate Mine} & 1h35 & 115k & 134 & 39.9M \\ 
			\midrule
			\makecell{Jordan} & \makecell{King Abdullah Reef} & 3h12 & 231k & 548 & 89.7M \\
			\midrule
			\makecell{Jordan} & \makecell{North Beach} & 0h27 & 32k & 202 & 39.4M\\
			\midrule
			\makecell{Jordan} & \makecell{Japanese Garden} & 2h3 & 147k & 198 & 26.3M\\
			\bottomrule
		\end{tabular}
		\vspace{-5pt}
		\label{table:videos}
	\end{table}

	\begin{figure}[hb!]
		\centering
		\includegraphics[width=0.49\linewidth]{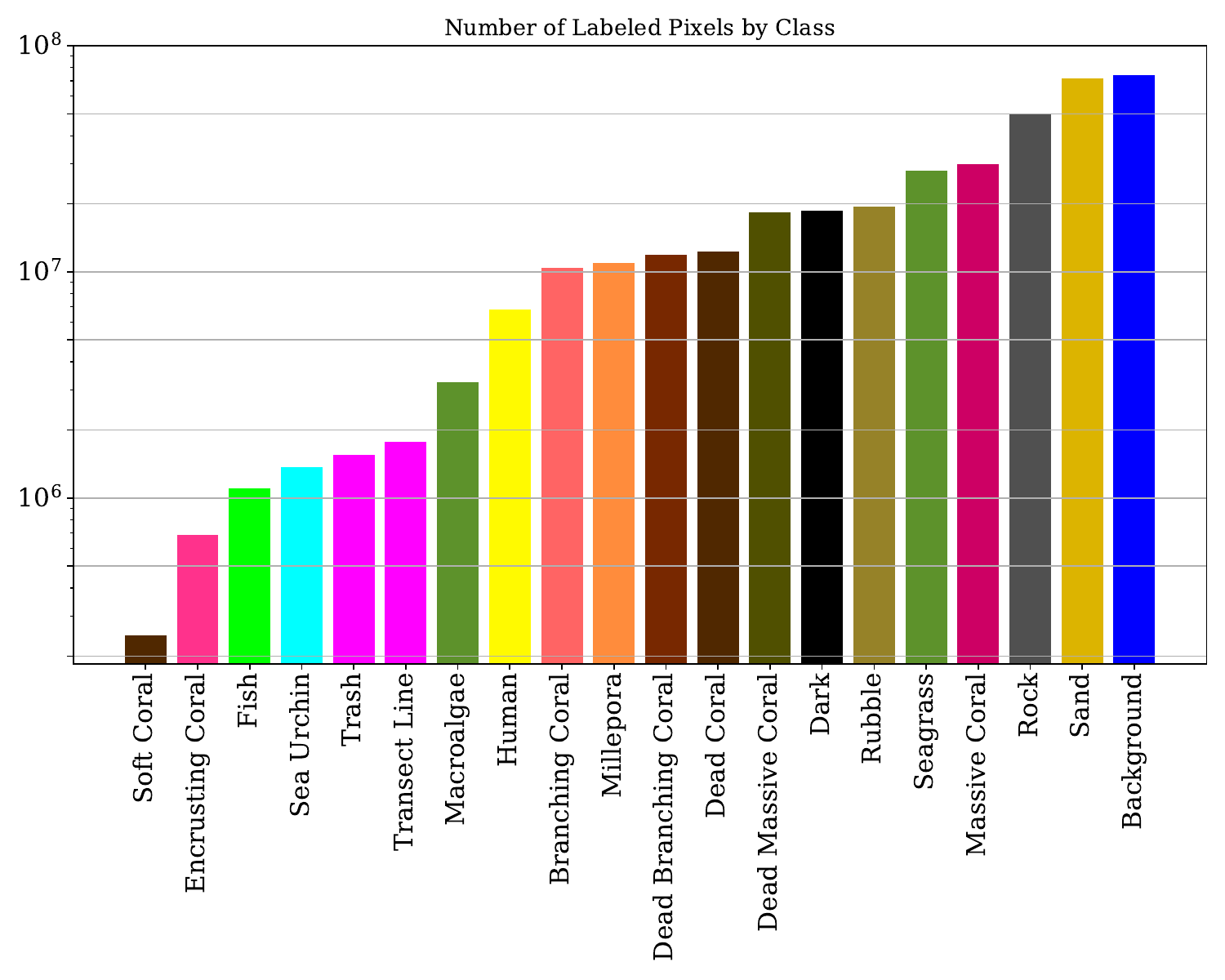}
		\includegraphics[width=0.49\linewidth]{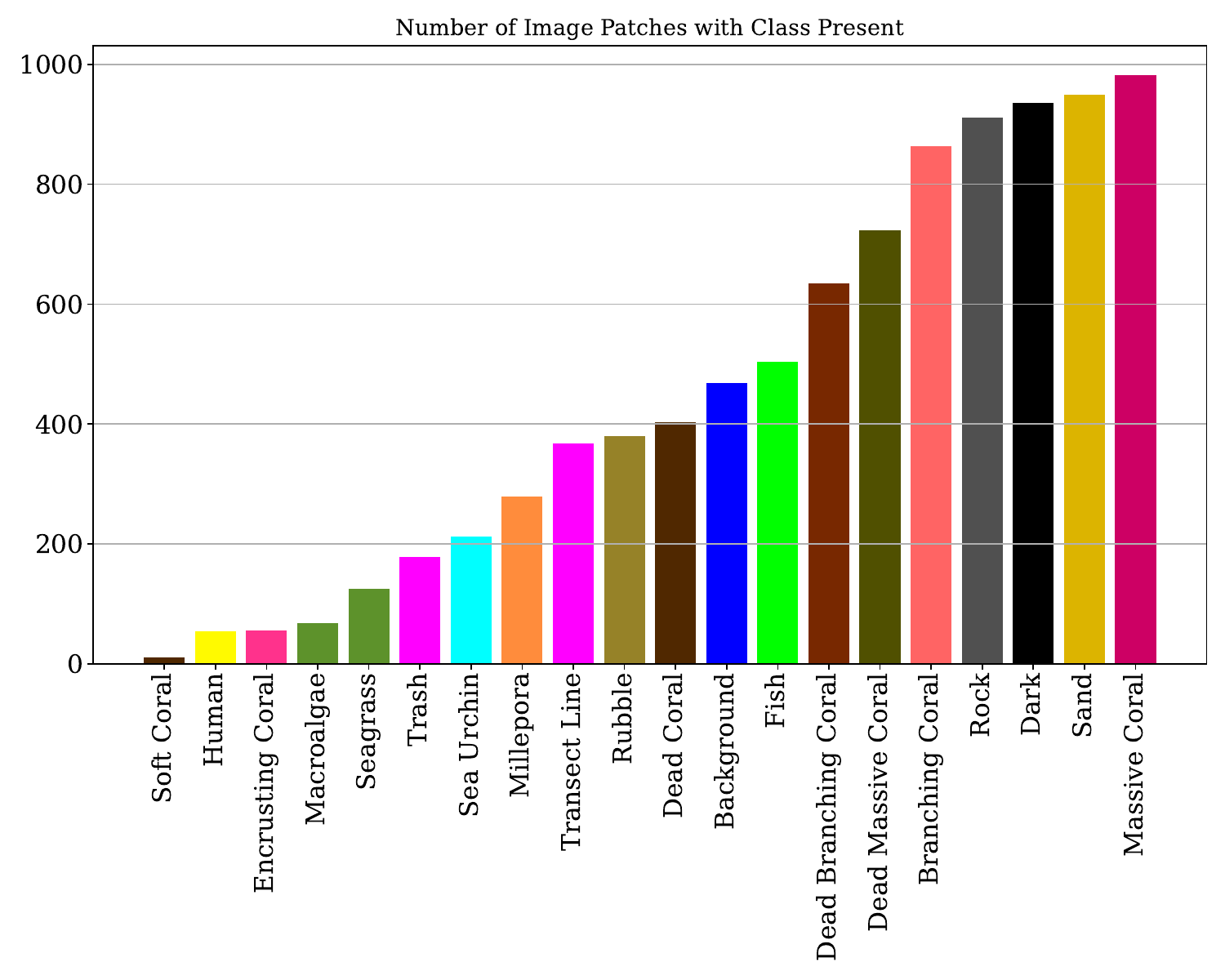}
		\vspace{-5pt}
		\caption{The distribution of labeled pixels by label class (left), and the number of image patches in which a label class is annotated (right).}
		\vspace{-5pt}
		\label{fig:labelclasses}
	\end{figure}

	\begin{figure}[hb!]
		\centering
		\includegraphics[width=0.99\linewidth]{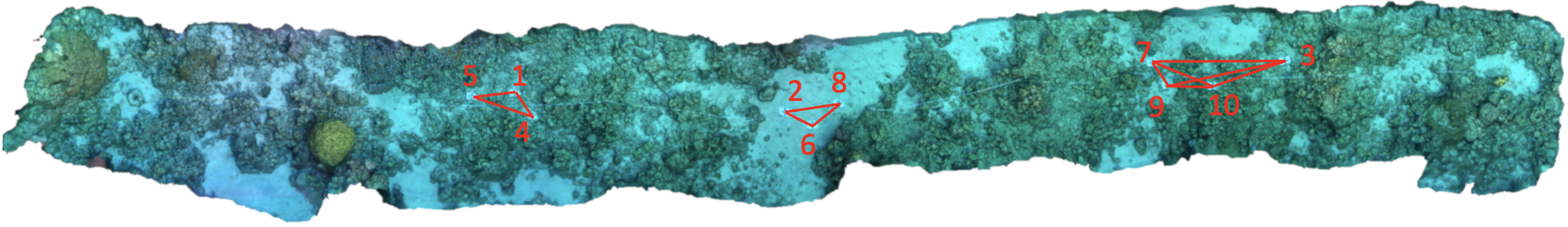}
		\caption{\new{Birds-eye view of the spatial evaluation transect (38 m in length) with ground truth markers placed. Red lines indicate the measured ground truth distances between markers that were measured by hand. Ortho-view made from high-resolution imagery with Agisoft Metashape.}}
		\label{fig:markers}
	\end{figure}

	\begin{figure}[hb!]
		\centering
		\includegraphics[width=0.99\linewidth]{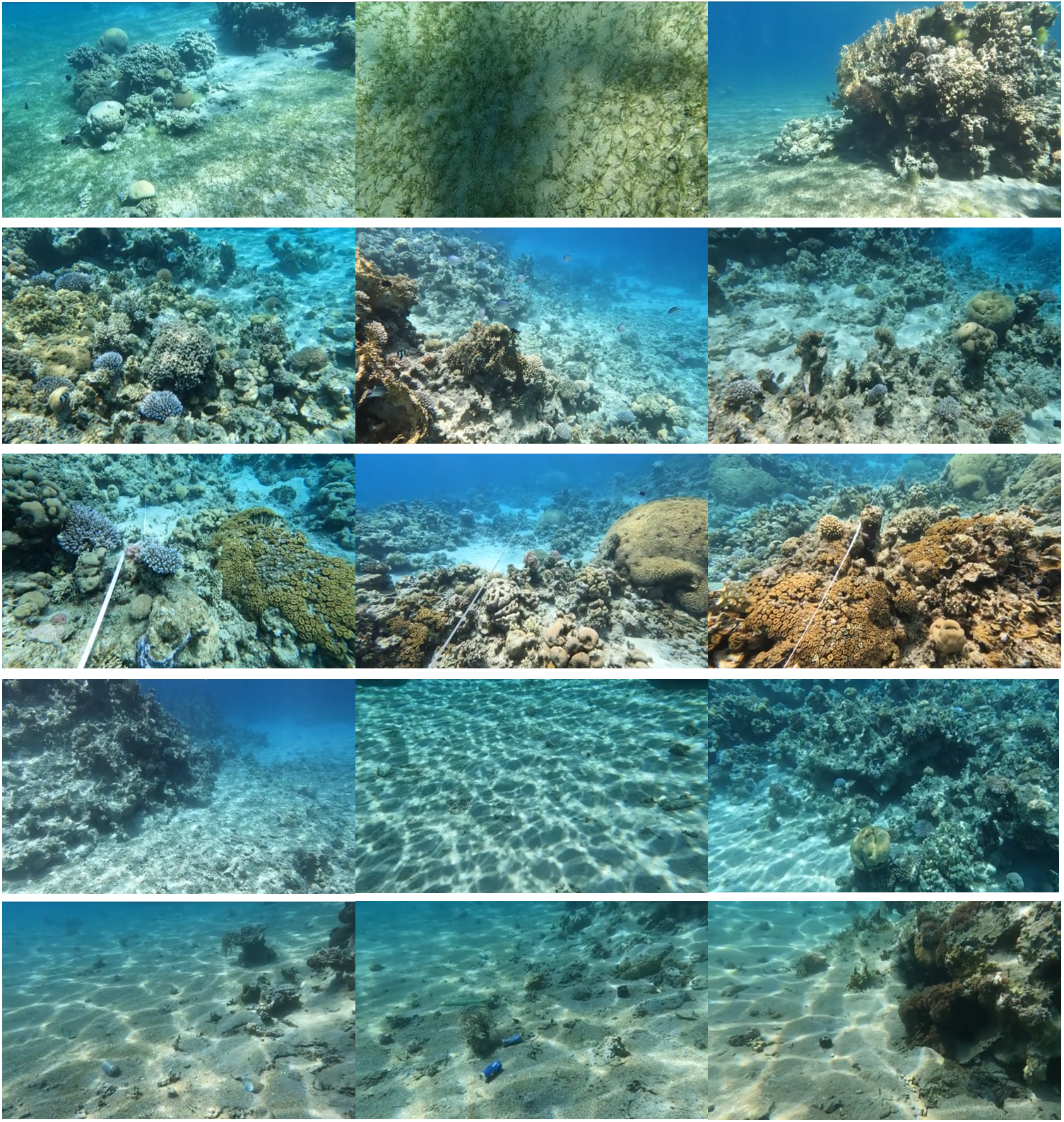}
		\caption{\new{Randomly selected video frames from the ego-motion video dataset, showing the diversity of the scenes: from patchy reefs on sea-grass covered seabeds in the South Phosphate Site (top row), over structurally complex reefs in the King Abdullah Reef (second row) and the Japanese Garden in Eilat (third row), scenes with extremely challenging caustics from the MSS Reef (fourth row), and the North Beach (bottom row), a site heavily impacted by humans, with visible litter.}}
		\label{fig:imageExamples}
	\end{figure}

	\clearpage

	\subsection*{Appendix: Implementation Details}

\begin{figure}[h]
		\centering
		\includegraphics[width=0.6\linewidth]{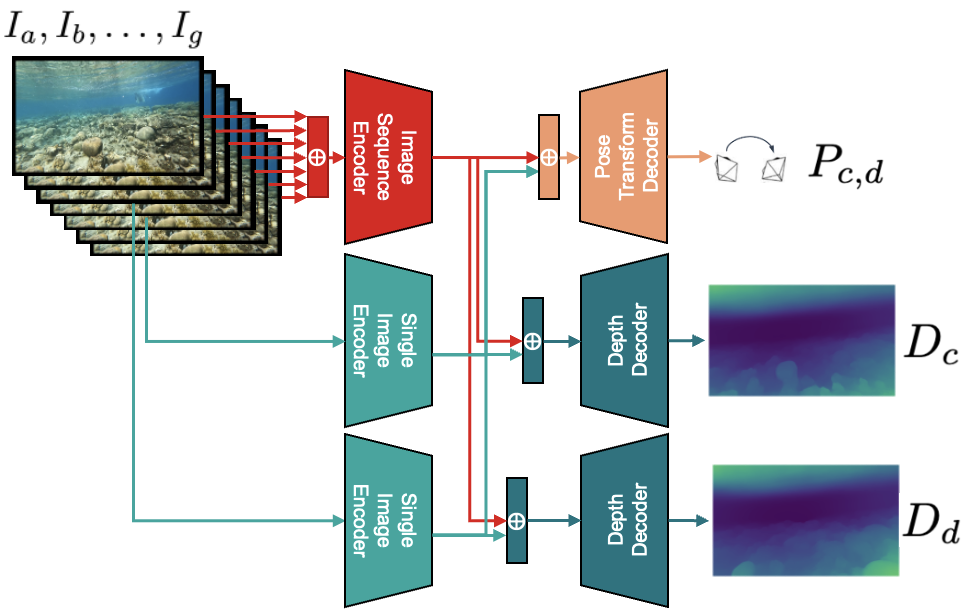}
		\caption{Schematic Figure showing the architecture of the depth \& pose estimation neural networks, which is a ResNet-34-encoder U-Net. One encoder encodes a sequence of seven images, the other encoder takes a single image as input. The features extracted by the encoders are concatenated along the channel dimension (denoted by $\oplus$). In this example, the calculation of the respective depths and camera pose transform of the third image $I_c$ and fourth image $I_d$ are highlighted.}
		\label{fig:architecture}
	\end{figure}

	Our implementation of a learning-based SfM system is based on the Scale Consistent SfM-Learner \citep{scaleconsistent}, which in turn is based on the seminar learned SfM system \citep{sfmlearner}. To combat the fact that some video frames exhibit strong blur due to the conditions under water and the low-cost camera setup, we slightly modify the implementation of the Scale Consistent SfM-Learner such that the neural network receives information about a larger sequence of images at a time instead of only one. This is realized by adding a second ResNet-34 image encoder, which takes seven images that are concatenated along the channel (color) dimensions. The encoded features of the image sequence are concatenated with features from a ResNet-34 encoder which takes one image at a time. The depth is computed from the concatenation of a single image's features with the sequence features, the pose is computed from the concatenation of two images' features with the sequence features. This neural network architecture change is illustrated in Figure \ref{fig:architecture}.
	
	Following this architecture change, we receive estimated depths and poses for all subsequent images in the image sequence (seven images at once). This means, that at point cloud creation time, the prediction acts as a sliding window, giving multiple depth estimates for each image, and multiple pose transform estimates for subsequent image pairs. Computing the variance of these predictions acts as a surrogate for uncertainty. In the point cloud creation process, the 35\% of pixels with the highest depth uncertainty are excluded from being added to the point cloud. As a postprocessing step, the 20\% of points in each point cloud with the highest depth uncertainty are removed.

\end{document}